\documentclass[letterpaper, 10 pt, journal, twoside]{ieeetran}
\IEEEoverridecommandlockouts

\pdfoutput=1
\usepackage{cite}
\usepackage{amsmath,amssymb,amsfonts}
\usepackage{algorithmic}
\usepackage{graphicx}
\usepackage{textcomp}
\usepackage{xcolor}
\usepackage{siunitx}
\usepackage{booktabs}
\usepackage{tabularx}
\usepackage{multirow}       
\usepackage{array} 
\usepackage{makecell}
\usepackage{amsmath}
\usepackage{lineno} 
\usepackage{caption}
\usepackage[font=footnotesize]{caption}
\usepackage{soul}
\usepackage{algorithm}
\usepackage{algorithmic}
\usepackage{subcaption}

\soulregister{\cite}7 
\soulregister{\citep}7 
\soulregister{\citet}7 
\soulregister{\ref}7 
\soulregister{\pageref}7 
\soulregister{\eqref}7 
\soulregister{\ang}7
\usepackage[caption=false,font=footnotesize,labelfont=rm,textfont=rm]{subfig}

\usepackage{tikz,xcolor}
\usepackage[hidelinks]{hyperref}

\definecolor{highlightcolor}{RGB}{255,0,0}

\definecolor{lime}{HTML}{A6CE39}
\DeclareRobustCommand{\orcidicon}{%
    \begin{tikzpicture}
    \draw[lime, fill=lime] (0,0) 
    circle [radius=0.16] 
    node[white] {{\fontfamily{qag}\selectfont \tiny ID}};    \draw[white, fill=white] (-0.0625,0.095) 
    circle [radius=0.007];    \end{tikzpicture}
    \hspace{-2mm}}
\foreach \x in {A, ..., Z}{%
    \expandafter\xdef\csname orcid\x\endcsname{\noexpand\href{https://orcid.org/\csname orcidauthor\x\endcsname}{\noexpand\orcidicon}}
    }

\hypersetup{
    colorlinks=False,
    linkcolor=blue, 
    urlcolor=black 
    citecolor=blue, 
    filecolor=black, 
    menucolor=black 
}

\def\BibTeX{{\rm B\kern-.05em{\sc i\kern-.025em b}\kern-.08em
    T\kern-.1667em\lower.7ex\hbox{E}\kern-.125emX}}

\makeatletter
\def\sf@counterlist{}
\makeatother

\usepackage{xpatch}

\makeatletter
\xpatchcmd{\IEEEbiography}{plus 1fil}{}{}{}
\xpatchcmd{\endIEEEbiography}{plus 1fil}{}{}{}
\xpatchcmd{\IEEEbiographynophoto}{plus 1fil}{}{}{}
\xpatchcmd{\endIEEEbiographynophoto}{plus 1fil}{}{}{}

\def\@IEEEBIOskipN{1.85\baselineskip}
\makeatother
  
\begin{document}

\title{Dual-Regime Hybrid Aerodynamic Modeling of Winged Blimps With Neural Mixing\\
\author{Xiaorui Wang, Hongwu Wang, Yue Fan, Hao Cheng, and Feitian Zhang\textsuperscript{*}}

\thanks{The authors are with the Robotics and Control Laboratory, School of Advanced Manufacturing and Robotics, and the State Key Laboratory of Turbulence and Complex Systems,  Peking University, Beijing, 100871, China (email: \href{mailto:jnswxr@stu.pku.edu.cn}{jnswxr@stu.pku.edu.cn}; \href{mailto:wanghongwu515@163.com}{wanghongwu515@163.com}; \href{mailto:2301213157@stu.pku.edu.cn}{2301213157@stu.pku.edu.cn}; \href{mailto:h-cheng@stu.pku.edu.cn}{h-cheng@stu.pku.edu.cn}; \href{mailto:feitian@pku.edu.cn}{feitian@pku.edu.cn}).

The dataset is publicly available at \url{https://github.com/ChaoxiWXR/Winged-Blimps-Dual-Regime-Hybrid-Aerodynamic-Modeling-Dataset}.
}

}

\maketitle
 \pagestyle{empty}  
\thispagestyle{empty} 

\begin{abstract}
Winged blimps operate across distinct aerodynamic regimes that cannot be adequately captured by a single model. At high speeds and small angles of attack, their dynamics exhibit strong coupling between lift and attitude, resembling fixed-wing aircraft behavior. At low speeds or large angles of attack, viscous effects and flow separation dominate, leading to drag-driven and damping-dominated dynamics. Accurately representing transitions between these regimes remains a fundamental challenge.
This paper presents a hybrid aerodynamic modeling framework that integrates a fixed-wing Aerodynamic Coupling Model (ACM) and a Generalized Drag Model (GDM) using a learned neural network mixer with explicit physics-based regularization. The mixer enables smooth transitions between regimes while retaining explicit, physics-based aerodynamic representation. Model parameters are identified through a structured three-phase pipeline tailored for hybrid aerodynamic modeling.
The proposed approach is validated on the RGBlimp platform through a large-scale experimental campaign comprising 1,320 real-world flight trajectories across 330 thruster and moving mass configurations, spanning a wide range of speeds and angles of attack. Experimental results demonstrate that the proposed hybrid model consistently outperforms single-model and predefined-mixer baselines, establishing a practical and robust aerodynamic modeling solution for winged blimps.
\end{abstract}


\section{Introduction}
\IEEEPARstart{R}{obotic} blimps represent a compelling class of aerial robots due to their combination of low energy consumption, long endurance, and intrinsic safety near people and infrastructure \cite{Tmech-2021-swing-reducing, Tmech-2021-scloud, TRO-RGBlimp-Q, RAL-2018-sense, RAL-env-sense-2024, IROS-2017-interact, ICRA-2019-model, Blimp-review-2024}. These characteristics make them particularly suitable for persistent monitoring \cite{RAL-2018-sense}, environmental sensing \cite{RAL-env-sense-2024}, and human-centered robotics tasks where prolonged flight and gentle interactions are essential \cite{IROS-2017-interact}. Accurate dynamical models are critical for such applications, as they enable reliable state estimation, trajectory planning, and control synthesis. However, predictive aerodynamic modeling for blimps is inherently challenging. The combination of large, flexible envelopes, low flight speeds, and coupled effects of buoyancy, propulsion, and ambient flow generates highly nonlinear forces and moments that are difficult to capture with a single global first principle model \cite{ICRA-2019-model, Blimp-review-2024}. This challenge becomes more pronounced in winged blimps, where additional lifting surfaces introduce multiple aerodynamic regimes with qualitatively distinct characteristics.

Winged blimps, exemplified by the RGBlimp platform \cite{TRO-RGBlimp-Q, RGBlimp-RAL}, operate across two qualitatively distinct aerodynamic regimes.  At moderate-to-high speeds and small angles of attack, wing-generated lift dominates and couples strongly with moments, producing dynamics analogous to fixed-wing aircraft
\cite{Fixed_wing_fig_AIAA, 2013-ACM-book, fixed-wing-TIM-2021}. In contrast, at low speeds or large angles of attack, viscous effects and flow separation dominate, producing primarily dissipative, drag-driven dynamics that resemble lighter-than-air vehicle behavior\cite{fluid-2023-Re, Re-2023-aerospace}. Unlike conventional fixed-wing or lighter-than-air aircraft, winged blimps routinely operate across both aerodynamic conditions due to their inherently slow cruise, large envelopes and agile maneuvers enabled by wing-generated lift \cite{NSTSMC-2023-model-likeAUV, ICARM-2024-blimp-drag-control}. Accurately modeling both regimes and their transitions within a unified framework remains a fundamental challenge.

\begin{figure*}[tbp]
\centerline{\includegraphics[width=1.0\linewidth]{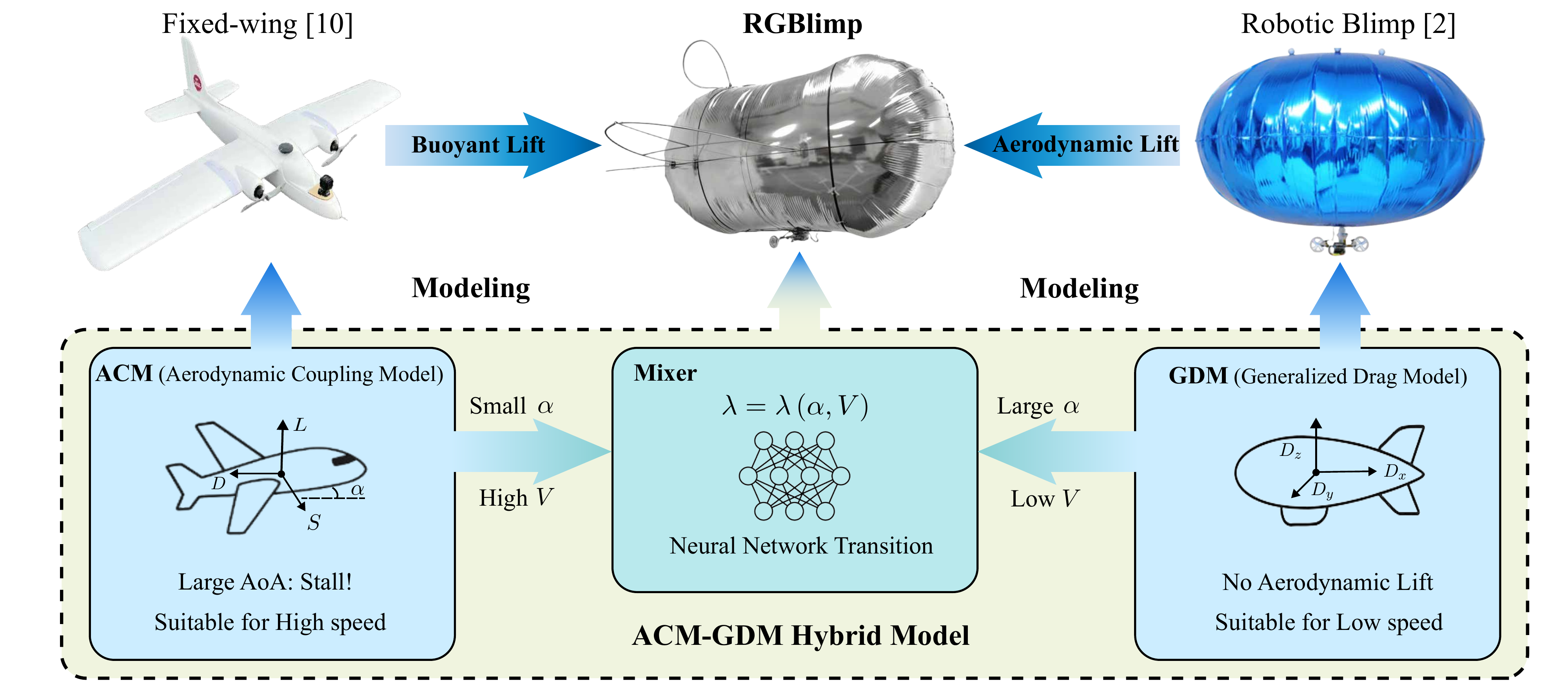}}
\caption{Overview of the ACM--GDM hybrid aerodynamic model for winged blimps. Using RGBlimp as an example, the Aerodynamic Coupling Model (ACM) captures lift-drag coupling and angle-dependent moments in the high-speed, small-angle-of-attack flight regime, while the Generalized Drag Model (GDM) represents drag-dominated, viscous effects in the low-speed, large-angle-of-attack operating regime. A neural-network mixer provides a smooth, physically regularized transition between the two regimes. The resulting hybrid model reproduces the dominant aerodynamic behaviors of winged blimps across their full operating envelopes, offering a unified and interpretable modeling framework.}
\label{fig-intro}
\end{figure*}

Existing modeling approaches typically target a single regime. Drag-centered models, often inspired by autonomous underwater vehicle (AUV) hydrodynamics\cite{fossen}, represent motion primarily through speed-dependent linear and quadratic damping terms \cite{ACC-model-likeAUV, CCC-2023-model-likeAUV}. While interpretable and effective in drag-dominated conditions \cite{SBlimp-IROS-2023-control, ICSL-2024-blimp-model-control}, these models fail to capture wing-generated lift and angle-dependent coupling. Fixed-wing style models accurately describe lift-drag-moment interactions at higher speeds and small angles of attack but lose fidelity at low speeds or post-stall conditions, where flow separation and viscous effects dominate \cite{stall-2015, high-alpha-2021, bluff-body-2022, fluid-2023-Re, Re-2023-aerospace}. Purely data-driven approaches fit these multi-regime dynamics empirically \cite{Neural-ODE, ABNODE}, yet they often compromise interpretability and generalization beyond the training data.

To address these challenges, this paper proposes a hybrid aerodynamic modeling framework, as illustrated in Fig.~\ref{fig-intro}, that integrates a fixed-wing \textbf{Aerodynamic Coupling Model (ACM)} for the lift-dominated regime and a \textbf{Generalized Drag Model (GDM)} for the drag-dominated regime. A lightweight neural network, trained with explicit physics-based regularization, smoothly blends the outputs of ACM and GDM across intermediate conditions. This approach ensures continuity across regimes while preserving interpretable, physics-based force decomposition. Model parameters are identified using a three-phase pipeline. Submodels are first calibrated in their dominant regimes, followed by training the transition network on intermediate flight data. 

The proposed framework is validated on the RGBlimp platform through a large-scale experimental campaign, comprising 1320 flight trajectories across 330 thruster and moving-mass configurations, spanning a wide range of speeds and angles of attack. Region-specific evaluation demonstrates that the hybrid model consistently outperforms single-model and fixed-switching baselines, offering improved predictive accuracy and robustness while retaining physical interpretability.

The main contributions of this paper are threefold.
First, we propose an ACM--GDM hybrid modeling framework for winged blimps, which unifies fixed-wing aerodynamic coupling with envelope-style drag formulations and enables accurate representation across dual flight regimes.
Second, we design a lightweight, physics-based neural transition mechanism and a three-phase identification pipeline that jointly recover submodel parameters and the regime transition map, ensuring smooth and interpretable regime blending.
Third, we collect a comprehensive experimental dataset and conduct extensive validation and ablation studies, demonstrating that the hybrid model consistently outperforms single-model and fixed-switching baselines, providing a practical and generalizable approach for winged blimp aerodynamic modeling.




\section{Motion Dynamics}\label{section-motion-dynamics}

\begin{figure*}[tbp]
\centerline{\includegraphics[width=1.0\linewidth]{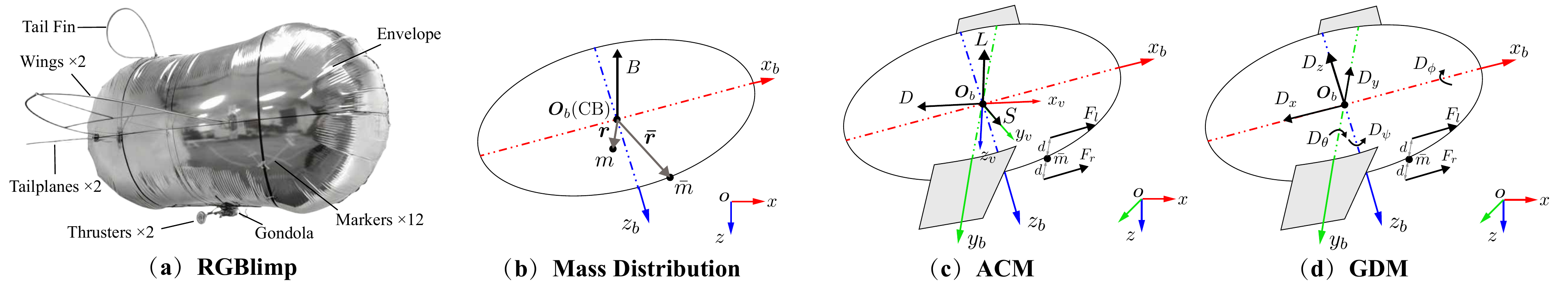}}
\caption{Illustration of the RGBlimp prototype and schematic representation of its ACM and GDM dynamic models. (a) RGBlimp prototype illustrating the winged envelope and suspended gondola with two thrusters. (b) Side-view diagram demonstrating the mass distribution. (c) Schematic of ACM, capturing lift--drag interactions and angle-dependent moments. (d) Schematic of GDM, representing drag-dominated, viscous effects.}
\label{fig-RGBlimp-and-Modeling}
\end{figure*}

The RGBlimp, as illustrated in Fig.~\ref{fig-RGBlimp-and-Modeling}(a), consists of a helium-filled winged envelope and a suspended gondola. The envelope carries a pair of fixed wings mounted symmetrically on its sides. The gondola houses two propulsive thrusters that generate forward thrust and differential thrust for yaw control. In addition, the gondola serves as a movable ballast for online adjustment of the system center of mass (CoM).

As illustrated in Fig.~\ref{fig-RGBlimp-and-Modeling}(b), the winged envelope is modeled as a rigid body. The inertial reference frame is denoted by $O$-$xyz$, and the body-fixed frame by $O_b$-$x_by_bz_b$ with its origin at the center of buoyancy (CoB). The envelope mass is $m_0$, and the center-of-gravity offset relative to the CoB is $\boldsymbol{r}_0$ in the body frame. The kinematics are
\begin{equation}
\boldsymbol{\dot{\eta}}=\boldsymbol{J}\left( \boldsymbol{\eta } \right) \boldsymbol{\nu }
\label{eq-kinematics}
\end{equation}
Here, $\boldsymbol{\eta }=\left[ x, y, z, \phi, \theta, \psi \right]^T$ represents the position and Euler angles in the inertial frame, and $\boldsymbol{\nu }=\left[ u, v, w, p, q, r \right]^T$ represents the translational and angular velocities expressed in the body frame. We further define $\boldsymbol{\eta}_1=\left[ x, y, z \right]^T$, $\boldsymbol{\eta}_2=\left[ \phi, \theta, \psi \right]^T$, $\boldsymbol{v}_b=\left[ u, v, w \right]^T$, and $\boldsymbol{\omega}_b=\left[ p, q, r \right]^T$. The Jacobian $\boldsymbol{J}\in \mathbb{R}^{6\times 6}$ is defined as
\begin{equation}
\boldsymbol{J}\left( \boldsymbol{\eta } \right)
=
\left[ \begin{matrix}
\boldsymbol{R}\left( \boldsymbol{\eta }_2 \right)& \mathbf{0}_{3\times 3}\\
\mathbf{0}_{3\times 3}& \boldsymbol{R}_{\omega}\left( \boldsymbol{\eta }_2 \right)
\end{matrix} \right]
\end{equation}
where $\boldsymbol{R}\left( \boldsymbol{\eta }_2 \right)\in \mathrm{SO}(3)$ maps the body frame to the inertial frame, and $\boldsymbol{R}_{\omega}\left( \boldsymbol{\eta }_2 \right)$ maps $\boldsymbol{\omega}_b$ to Euler-angle rates.

The envelope rigid-body dynamics are
\begin{equation}
\boldsymbol{M}_0\boldsymbol{\dot{\nu}}+\boldsymbol{C}_0\left( \boldsymbol{\nu } \right) \boldsymbol{\nu }+\boldsymbol{g}_0\left( \boldsymbol{\eta } \right) =\boldsymbol{F}_{\text{aero}} + \boldsymbol{F}_{\text{add}}
\end{equation}
where $\boldsymbol{F}_{\text{aero}}$ denotes the velocity-dependent aerodynamic loads and $\boldsymbol{F}_{\text{add}}$ denotes added-mass effects. The standard rigid-body terms are
\begin{equation}
\boldsymbol{M}_0=\left[ \begin{matrix}
m_0\mathbf{1}_{3\times 3}& -m_0\mathbf{S}\left( \boldsymbol{r}_0 \right)\\
m_0\mathbf{S}\left( \boldsymbol{r}_0 \right)& \boldsymbol{I}_0
\end{matrix} \right]
\end{equation}
\begin{equation}
\boldsymbol{C}_0\left( \boldsymbol{\nu } \right) =\left[ \begin{matrix}
m_0\mathbf{S}\left( \boldsymbol{\omega }_b \right)& -m_0\mathbf{S}\left( \boldsymbol{\omega }_b \right) \mathbf{S}\left( \boldsymbol{r}_0 \right)\\
-m_0\mathbf{S}\left( \boldsymbol{\omega }_b \right) \mathbf{S}\left( \boldsymbol{r}_0 \right)& -\mathbf{S}\left( \boldsymbol{I}_0\boldsymbol{\omega }_b \right)
\end{matrix} \right]
\end{equation}
\begin{equation}
\boldsymbol{g}_0\left( \boldsymbol{\eta } \right) =\left[ \begin{array}{c}
\left( m_0g-B \right) \boldsymbol{R}^{\text{T}}\boldsymbol{k}\\
m_0g\mathbf{S}\left( \boldsymbol{r}_0 \right) \boldsymbol{R}^{\text{T}}\boldsymbol{k}
\end{array} \right]
\end{equation}
Here, $\boldsymbol{I}_0$ is the inertia tensor of the envelope, $g$ is the gravitational acceleration, $B$ represents the net buoyant force, $\boldsymbol{k}=\left[ 0,0,1 \right]^T$ is the unit vector along the inertial $z$-axis, and $\boldsymbol{S}(\cdot)$ denotes the skew-symmetric operator.

The gondola is modeled as a point mass $\bar{m}$ located at $\boldsymbol{\bar{r}}=\left[ \bar{r}_x, \bar{r}_y, \bar{r}_z \right]^T$ in the body frame. The thruster-generated control wrench is
\begin{equation}
\boldsymbol{\tau }=\left[ \begin{array}{c}
\left( F_l+F_r \right) \boldsymbol{i}_b\\
\left( F_l+F_r \right) \bar{r}_z\boldsymbol{j}_b+\left( F_l-F_r \right) d\boldsymbol{k}_b
\end{array} \right]
\end{equation}
where $F_l$ and $F_r$ denote the thrust force magnitudes of the left and right propellers, respectively. $\boldsymbol{i}_b = \left[ 1,0,0 \right]^T$, $\boldsymbol{j}_b = \left[ 0,1,0 \right]^T$, and $\boldsymbol{k}_b = \left[ 0,0,1 \right]^T$ are unit vectors along the $O_bx_b$, $O_by_b$, and $O_bz_b$ axes, respectively. $d$ represents the lateral offset from each propeller to the $O_bx_b$ axis.

Following D’Alembert’s principle, the gondola exerts a reaction force on the envelope
\begin{equation}
\boldsymbol{\bar{f}}=\bar{m}g\boldsymbol{R}^{\mathrm{T}}\boldsymbol{k}-\bar{m}\boldsymbol{\dot{\bar{v}}}
\end{equation}
where the gondola acceleration in the body frame is
\begin{equation}
\boldsymbol{\dot{\bar{v}}}
=
\dot{\boldsymbol{v}}_b
+\dot{\boldsymbol{\omega}}_b\times\bar{\boldsymbol{r}}
+\boldsymbol{\omega}_b\times\!\left(\boldsymbol{\omega}_b\times\bar{\boldsymbol{r}}\right)
+2\,\boldsymbol{\omega}_b\times\dot{\bar{\boldsymbol{r}}}
+\ddot{\bar{\boldsymbol{r}}}
\label{eq-gondola-acc}
\end{equation}
The corresponding reaction moment is $\boldsymbol{\bar{t}}=\bar{\boldsymbol{r}}\times \boldsymbol{\bar{f}}$.

Let the total mass be $m = m_0 + \bar{m}$ and define the overall CoM offset $\boldsymbol{r}$ such that $m\boldsymbol{r} = m_0\boldsymbol{r}_0 + \bar{m}\boldsymbol{\bar{r}}$. The full rigid-body dynamics are
\begin{equation}
\boldsymbol{M}_{RB}\boldsymbol{\dot{\nu}}+\boldsymbol{C}_{RB}\left( \boldsymbol{\nu } \right) \boldsymbol{\nu }+\boldsymbol{g}\left( \boldsymbol{\eta } \right) =\boldsymbol{\tau }+\boldsymbol{F}_{\text{aero}} + \boldsymbol{F}_{\text{add}} +\boldsymbol{\bar{F}}
\end{equation}
where
\begin{equation}
\boldsymbol{M}_{RB}=\left[ \begin{matrix}
m\mathbf{1}_{3\times 3}& -m\mathbf{S}\left( \boldsymbol{r} \right)\\
m\mathbf{S}\left( \boldsymbol{r} \right)& \boldsymbol{I}_0-\bar{m}\mathbf{S}\left( \boldsymbol{\bar{r}} \right) \mathbf{S}\left( \boldsymbol{\bar{r}} \right)
\end{matrix} \right]
\end{equation}
\begin{equation}
\boldsymbol{C}_{RB}\left( \boldsymbol{\nu } \right) =\left[ \begin{matrix}
m\mathbf{S}\left( \boldsymbol{\omega }_b \right)& -m\mathbf{S}\left( \boldsymbol{\omega }_b \right) \mathbf{S}\left( \boldsymbol{r} \right)\\
-m\mathbf{S}\left( \boldsymbol{\omega }_b \right) \mathbf{S}\left( \boldsymbol{r} \right)& -\mathbf{S}\left( \left( \boldsymbol{I}_0-\bar{m}\mathbf{S}\left( \boldsymbol{\bar{r}} \right) \mathbf{S}\left( \boldsymbol{\bar{r}} \right) \right) \boldsymbol{\omega }_b \right)
\end{matrix} \right]
\end{equation}
\begin{equation}
\boldsymbol{g}\left( \boldsymbol{\eta } \right) =\left[ \begin{array}{c}
\left( mg-B \right) \boldsymbol{R}^{\text{T}}\boldsymbol{k}\\
mg\mathbf{S}\left( \boldsymbol{r} \right) \boldsymbol{R}^{\text{T}}\boldsymbol{k}
\end{array} \right]
\end{equation}
\begin{equation}
\boldsymbol{\bar{F}}=\left[ \begin{array}{c}
-\bar{m}\boldsymbol{\ddot{\bar{r}}}+2\bar{m}\mathbf{S}\left( \boldsymbol{\dot{\bar{r}}} \right) \boldsymbol{\omega }_b\\
-\bar{m}\mathbf{S}\left( \boldsymbol{\bar{r}} \right) \boldsymbol{\ddot{\bar{r}}}+2\bar{m}\mathbf{S}\left( \boldsymbol{\bar{r}} \right) \mathbf{S}\left( \boldsymbol{\dot{\bar{r}}} \right) \boldsymbol{\omega }_b
\end{array} \right]
\end{equation}

Finally, the added-mass effects are incorporated as
\begin{equation}
\begin{aligned}
&\left( \boldsymbol{M}_{RB}+\boldsymbol{M}_A \right) \boldsymbol{\dot{\nu}}+\left( \boldsymbol{C}_{RB}\left( \boldsymbol{\nu } \right) +\boldsymbol{C}_A\left( \boldsymbol{\nu } \right) \right) \boldsymbol{\nu }
+\boldsymbol{g}\left( \boldsymbol{\eta } \right) \\
&\qquad =\boldsymbol{\tau }+\boldsymbol{F}_{\text{aero}} +\boldsymbol{\bar{F}}
\end{aligned}
\end{equation}

Due to the geometric symmetry of the blimp, the added-mass matrix is approximated as diagonal, i.e.
\begin{equation}
\boldsymbol{M}_A=\text{diag}\left( X_{\dot{u}},Y_{\dot{v}},Z_{\dot{w}},K_{\dot{p}},M_{\dot{q}},N_{\dot{r}} \right)
\end{equation}
while $\boldsymbol{C}_A\left( \boldsymbol{\nu } \right)$ accounts for the corresponding velocity-dependent coupling terms induced by the added-mass effects.

The remaining term, $\boldsymbol{F}_{\text{aero}}$, captures velocity-dependent aerodynamic forces and moments. In this work, $\boldsymbol{F}_{\text{aero}}$ is modeled using an ACM--GDM hybrid formulation. The detailed structure of this aerodynamic model is presented in the subsequent sections.

\section{Hybrid Aerodynamic Modeling}\label{section-Hybrid-Aerodynamic-Modeling}
\subsection{Aerodynamic Coupling Model}
When operating at moderate-to-high forward speeds and small angles of attack, the blimp's wings generate non-negligible aerodynamic forces and moments. In this flight regime, the aerodynamic behavior of the RGBlimp resembles that of a fixed-wing glider in which lift--drag coupling and angle-dependent aerodynamic coefficients dominate the dynamics. We refer to this regime and its corresponding formulation as the Aerodynamic Coupling Model (ACM).

To describe the ACM, we define the velocity reference frame $O_b$-$x_vy_vz_v$. The $O_bx_v$ axis is aligned with the instantaneous velocity vector, while the $O_bz_v$ axis lies in the blimp's plane of symmetry and is perpendicular to $O_bx_v$, as illustrated in Fig.~\ref{fig-RGBlimp-and-Modeling}(c). $\boldsymbol{R}_{v}^{b}$ denotes the rotation from the velocity frame to the body-fixed frame, i.e.,
\begin{equation}
    \boldsymbol{R}_{v}^{b}=\left[ \begin{matrix}
	\cos \alpha \cos \beta&		-\cos \alpha \sin \beta&		-\sin \alpha\\
	\sin \beta&		\cos \beta&		0\\
	\sin \alpha \cos \beta&		-\sin \alpha \sin \beta&		\cos \alpha\\
\end{matrix} \right]
\end{equation}
Here, $\alpha=\arctan\left(w/u\right)$ denotes the angle of attack, $\beta=\arcsin\left(v/V\right)$ denotes the sideslip angle, and $V=\sqrt{u^2+v^2+w^2}$ is the velocity magnitude.

In the ACM, aerodynamic forces and moments are first expressed in the velocity frame and subsequently mapped to the body frame as
\begin{equation}
    \boldsymbol{F}_{\text{aero}}^{\text{ACM}}
    =
    \boldsymbol{R}_{v}^{b}
    \left[
    \begin{matrix}
        -D & S & -L & M_1 & M_2 & M_3
    \end{matrix}
    \right]^{\text{T}}
\end{equation}
where $D$, $S$, and $L$ denote the drag, side force, and lift, respectively, and $M_1$, $M_2$, and $M_3$ represent the aerodynamic moments about the roll, pitch, and yaw axes.

Following the formulation in \cite{2013-ACM-book}, the aerodynamic loads in the velocity frame are parameterized by angle-dependent coefficients scaled by dynamic pressure, together with linear rotational damping terms for moments. For example, the drag and pitch moment are modeled as $D=\tfrac{1}{2}\rho V^2 A\,C_D(\alpha,\beta)$ and $M_2=\tfrac{1}{2}\rho V^2 A\,C_{M_2}(\alpha,\beta)+K_2 q$, and the remaining force and moment components are parameterized in the same manner. Here, $\rho$ denotes the air density and $A$ is the reference area. $K_1$, $K_2$, and $K_3$ are the rotational damping coefficients. The coefficient functions $C_{\Box}\!\left(\alpha,\beta\right)$ capture the coupled dependence of aerodynamic forces and moments on the angle of attack and sideslip angle, and they are identified from experimental or simulation data using low-order polynomial parameterizations in $\alpha$ and $\beta$ \cite{2013-ACM-book}.

\subsection{Generalized Drag Model}
When operating at sufficiently large angles of attack, the RGBlimp experiences a stall where the lift decreases sharply. Separated flow around the wings and adjacent envelope surfaces generates aerodynamic forces resembling bluff-body behavior. Furthermore, at low airspeeds, the Reynolds number decreases and viscous effects become dominant, requiring both pressure and viscous contributions to be considered \cite{2022-Re, aerospace-2024-Re}. To model the aerodynamics in this drag-dominated regime, we employ a Generalized Drag Model (GDM), as illustrated in Fig.~\ref{fig-RGBlimp-and-Modeling}(d). Under the GDM, the aerodynamic force $\boldsymbol{F}_{\text{aero}}$ is written as
\begin{equation}
    \boldsymbol{F}_{\text{aero}}^{\text{GDM}}
    =
    -\boldsymbol{D}\left( \boldsymbol{\nu } \right) \boldsymbol{\nu }
    =
    -\left( \boldsymbol{D}_L +\boldsymbol{D}_Q \odot \left| \boldsymbol{\nu } \right| \right) \boldsymbol{\nu }
\end{equation}
where
\begin{equation}
    \boldsymbol{D}_L=\text{diag}\left( X_u,Y_v,Z_w,K_p,M_q,N_r \right)
\end{equation}
\begin{equation}
    \boldsymbol{D}_Q=\text{diag}\left( X_{u\left| u \right|},Y_{v\left| v \right|},Z_{w\left| w \right|},K_{p\left| p \right|},M_{q\left| q \right|},N_{r\left| r \right|} \right)
\end{equation}
denote the diagonal matrices of linear and quadratic drag coefficients, respectively. The linear terms capture velocity-proportional damping effects that become significant at low Reynolds numbers, while the quadratic terms capture drag components that scale with dynamic pressure. Together, these terms approximate the total aerodynamic resistance acting against the local flow. Notably, the GDM does not explicitly model lift or lift-induced moments.

In summary, the GDM is applied in low-speed and high--angle-of-attack regimes, where stall and enhanced viscous effects make the dynamics strongly drag-dominated. In contrast, the ACM describes high-speed, small--angle-of-attack regimes, in which the wings generate significant lift and lift--drag coupling, and the aerodynamics are more appropriately represented in the velocity frame with explicit angle-of-attack and sideslip dependencies.

\subsection{Hybrid Model With Neural Mixing}
To accurately capture the blimp dynamics across a broad range of operating conditions, encompassing variations in speed, angle of attack, and the corresponding aerodynamic regimes, we formulate a unified ACM–GDM hybrid aerodynamic model. The complete hybrid model is expressed as
\begin{equation}
\begin{aligned}
\left( \boldsymbol{M}_{RB}+ \boldsymbol{M}_A \right) \boldsymbol{\dot{\nu}}+\left( \boldsymbol{C}_{RB}\left( \boldsymbol{\nu } \right) +\boldsymbol{C}_A\left( \boldsymbol{\nu } \right) \right) \boldsymbol{\nu }
+\boldsymbol{g}\left( \boldsymbol{\eta } \right)\\ =\boldsymbol{\tau }+\left( 1-\lambda \right) \boldsymbol{F}_{\text{aero}}^{\text{ACM}}+\lambda \boldsymbol{F}_{\text{aero}}^{\text{GDM}}+\boldsymbol{\bar{F}}
\end{aligned}
\label{eq-hybrid}
\end{equation}
Here, $\lambda$ is a mixing coefficient that regulates the relative contributions of the two aerodynamic models. Specifically, $\lambda=0$ recovers the ACM regime, while $\lambda=1$ corresponds to the GDM regime. To avoid discontinuities in the aerodynamic forces and moments, a transition region is defined in which $\lambda$ varies smoothly as a function of the angle of attack $\alpha$ and the velocity magnitude $V$. This functional relationship is approximated using a neural network, i.e.,
\begin{equation}
\lambda =\lambda \left( \alpha, V ;\boldsymbol{\xi } \right) 
\end{equation}
where $\boldsymbol{\xi}$ denotes the network parameters. The resulting ACM--GDM hybrid model provides a unified aerodynamic formulation that remains valid across heterogeneous flight conditions, while the learned transition function ensures smooth blending between regimes and maintains model consistency and accuracy throughout the entire operation of the blimp.

\section{Hybrid Model Identification}\label{section-Hybrid-model-identification}
\subsection{Identification Problem Formulation}
This section presents the parameter identification strategy for the proposed hybrid aerodynamic model. For conciseness, the full ACM–GDM hybrid dynamics of the blimp, defined in Eq.~\eqref{eq-kinematics} and Eq.~\eqref{eq-hybrid}, are rewritten in a more compact form as
\begin{equation}
\boldsymbol{\dot{x}}=\boldsymbol{h}\left( \boldsymbol{x,u;\varphi ,}\lambda \left( \alpha,V ;\boldsymbol{\xi } \right) \right) 
\label{eq-ODE}
\end{equation}
Here, $\boldsymbol{h}\left(\cdot\right)$ denotes the hybrid nonlinear dynamics, $\boldsymbol{x}=\left[\boldsymbol{\eta}^{\text{T}},\boldsymbol{\nu}^{\text{T}}\right]^{\text{T}}$ is the system state, and $\boldsymbol{u}=\left[F_l,F_r,\boldsymbol{\bar{r}}\right]^{\text{T}}$ is the control input. To reduce identification complexity and isolate aerodynamic effects, the gondola position is assumed constant within each experimental trial, i.e., $\boldsymbol{\dot{\bar{r}}}=0$. The vector $\boldsymbol{\varphi}$ represents the physical parameters of the blimp obtained from first principles modeling, while $\boldsymbol{\xi}$ denotes the parameters associated with the transition neural network.

To quantify model accuracy and enable iterative parameter updates, we adopt a loss function, inspired by \cite{TRO-2024-UKF} and \cite{KNODE-MPC-2022}, based on the discrepancy between the model’s forward prediction and the measured ground truth, defined as
\begin{equation}
\mathcal{L}_\text{model}=\frac{1}{N}\sum_{i=1}^N{\frac{1}{n}\lVert \boldsymbol{W}\left( \boldsymbol{x}_p\left( t_i \right) -\boldsymbol{x}\left( t_i \right) \right) \rVert _{2}^{2}}
\label{eq-loss}
\end{equation}
Here, $N$ denotes the number of data samples, $n$ is the dimension of the system state, and $\boldsymbol{W}\in \mathbb{R}^n\times \mathbb{R}^n$ is a diagonal weighting matrix reflecting the relative importance of different state components. $\boldsymbol{x}\left( t_i \right)$ represents the measured state at time $t_i$, while $ \boldsymbol{x}_p\left( t_i \right)$ denotes the corresponding model prediction. 
The predicted state $\boldsymbol{x}_p$ is generated according to the active aerodynamic regime. In the pure ACM regime, the forward propagation is represented as $ \boldsymbol{\bar{x}}$, i.e.,
\begin{equation}
    \boldsymbol{\bar{x}}\left( t_{i+1} \right) =\boldsymbol{x}\left( t_i \right) +\int_{t_i}^{t_{i+1}}{\boldsymbol{h}\left( \boldsymbol{x,u;\varphi ,}\lambda \right) _{\left| \lambda =0 \right.}\text{d}t}
\label{eq-X1}
\end{equation}
Whereas in the pure GDM regime it is denoted as $\boldsymbol{\tilde{x}}$ and given by
\begin{equation}
    \boldsymbol{\tilde{x}}\left( t_{i+1} \right) =\boldsymbol{x}\left( t_i \right) +\int_{t_i}^{t_{i+1}}{\boldsymbol{h}\left( \boldsymbol{x,u;\varphi ,}\lambda \right) _{\left| \lambda =1 \right.}\text{d}t}
\label{eq-X2}
\end{equation}
Within the transition region, the prediction, represented as $\boldsymbol{\hat{x}}$, employs the hybrid formulation with a mixing coefficient, i.e.,
\begin{equation}
    \boldsymbol{\hat{x}}\left( t_{i+1} \right) =\boldsymbol{x}\left( t_i \right) +\int_{t_i}^{t_{i+1}}{\boldsymbol{h}\left( \boldsymbol{x,u;\varphi ,}\lambda \left( \alpha,V ;\boldsymbol{\xi } \right) \right)\text{d}t} 
\label{eq-X3}
\end{equation}

\subsection{Neural Mixing}
In the transition region, we model the blending coefficient $\lambda$ using a feedforward neural network that takes the angle of attack $\alpha$ and the speed $V$ as inputs and outputs a continuous scalar in $\left[0,1\right]$, which determines the relative contributions of ACM and GDM. In addition to the model prediction loss $\mathcal{L}_\text{model}$, we introduce a set of physics-based regularizers that enforce physically meaningful transition behaviors.

First, \textbf{anchor-point regularization} is employed to enforce the network to produce the desired limits in the pure-regime regions. Specifically, for anchor points inside the ACM region, we require $\lambda \approx 0$, whereas for anchors inside the GDM region, we require $\lambda \approx 1$. The corresponding anchor loss is defined as the summation of squared errors over all the anchors, i.e.,
\begin{equation}
    \mathcal{L}_{\text{anchor}}=\sum_{i\in \mathcal{A}_{\text{ACM}}}{\left( \lambda \left( \alpha_i,V _i \right) -0 \right) ^2+\sum_{j\in \mathcal{A}_{\text{GDM}}}{\left( \lambda \left( \alpha_j,V _j \right) -1 \right) ^2}}
\end{equation}
where $\mathcal{A}_{\text{ACM}}$ and $\mathcal{A}_{\text{GDM}}$ denote the selected anchor sets.

Second, to preserve \textbf{physical monotonicity}, we penalize violations of the expected trends of $\lambda$ with respect to its inputs. Specifically, the contribution of the drag-dominated GDM is expected to increase with the angle of attack and decrease with airspeed. This behavior is enforced by penalizing positive gradients with respect to $\alpha$ and negative gradients with respect to $V$, yielding
\begin{equation}
    \mathcal{L}_{\text{mono}}=\sum_{\alpha ,V\in \mathcal{P}}{\max \left( 0,\partial _{\alpha}\lambda \right) ^2+\sum_{\alpha ,V\in \mathcal{P}}{\max \left( 0,-\partial _V\lambda \right) ^2}}
\end{equation}
Here, $\partial_{\alpha}\lambda$ and $\partial_{V}\lambda$ denote the partial derivatives of $\lambda$ with respect to $\alpha$ and $V$, respectively. In practice, these derivatives are approximated via finite differences over a predefined set of grid points $\mathcal{P}$ in the $\left(\alpha, V\right)$ domain.

Third, a \textbf{smoothness regularization} term is introduced to avoid spurious oscillations and to ensure a physically plausible and slowly varying transition surface, i.e.,
\begin{equation}
    \mathcal{L}_{\text{smooth}}=\sum_{\alpha ,V\in \mathcal{P}}{\left[ \left( \partial _{\alpha}\lambda \right) ^2 +\left( \partial _{V}\lambda \right) ^2 \right]}
\end{equation}

The overall loss used to train the transition network is  then defined as a weighted sum of the above terms, given by
\begin{equation}
\mathcal{L}_{\text{total}}=\mathcal{L}_{\text{model}}+w_{\text{anchor}}\mathcal{L}_{\text{anchor}}+w_{\text{mono}}\mathcal{L}_{\text{mono}}+w_{\text{smooth}}\mathcal{L}_{\text{smooth}}
\label{Loss_total}
\end{equation}
with scalar weights $w_{\text{anchor}}$, $w_{\text{mono}}$, and $w_{\text{smooth}}$ chosen to balance prediction accuracy against physical regularization. These constraints ensure that $\lambda$ attains the correct regime limits, varies monotonically with the key physical inputs, and transitions smoothly between regimes, yielding a physically consistent and numerically stable blending of the ACM and GDM.

\subsection{Three-phase Training}
\label{Section-Three-phase Training}

\begin{figure*}[tbp]
\centerline{\includegraphics[width=1.0\linewidth]{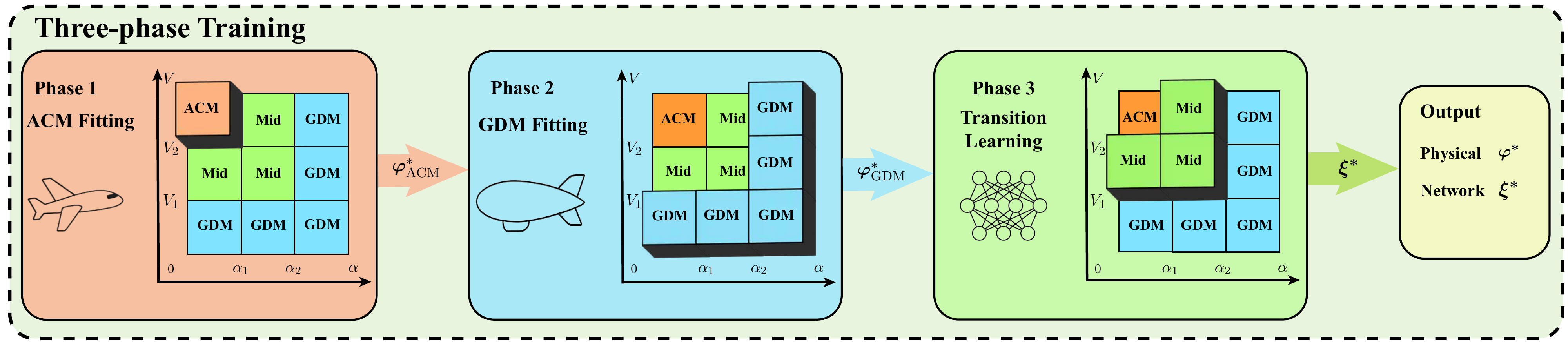}}
\caption{Three-phase identification pipeline of the ACM–GDM hybrid model. The flight envelope is partitioned by $\alpha_1$, $\alpha_2$, $V_1$, $V_2$ into ACM, GDM, and transition regions. Phase 1 sets the mixing coefficient $\lambda=0$ and uses ACM region data to identify ACM parameters $\boldsymbol{\varphi }_{\text{ACM}}^{*}$. Phase 2 sets $\lambda=1$ and uses GDM region data to identify GDM parameters $\boldsymbol{\varphi }_{\text{GDM}}^{*}$, yielding the complete set of physical parameters $\boldsymbol{\varphi }^*$. Phase 3 fixes $\boldsymbol{\varphi } = \boldsymbol{\varphi }^*$ and trains the neural network $\lambda =\lambda \left( \alpha,V ;\boldsymbol{\xi } \right) $ on transition-region data to obtain the network parameters $\boldsymbol{\xi^*}$. The procedure produces the identified physical parameters $\boldsymbol{\varphi }^*$ and the learned transition parameters $\boldsymbol{\xi^*}$.}
\label{fig-train-step}
\end{figure*}

As illustrated in Fig.~\ref{fig-train-step}, we propose a three-phase training strategy to identify the proposed hybrid dynamics model. The procedure explicitly exploits the separation of aerodynamic regimes to improve identifiability, numerical stability, and physical consistency. First, the dataset is partitioned in the $\left(\alpha, V\right)$ plane using four threshold parameters $\alpha_1$, $\alpha_2$, $V_1$, and $V_2$. Concretely, the ACM region is defined as the set of states with $\alpha < \alpha_1$ and $V > V_2$, corresponding to high-speed, small-angle-of-attack conditions. The transition region is constructed as an L-shaped band formed by states satisfying either $\alpha < \alpha_1$ with $V \in \left[V_1, V_2\right]$ or $\alpha \in \left[\alpha_1, \alpha_2\right]$ with $V > V_1$. The remaining portion of the plane is assigned to the GDM region, which captures low-speed and high-angle-of-attack conditions. Based on this partition, the collected data are split into three subsets $\left(\boldsymbol{X}_1, \boldsymbol{U}_1\right)$, $\left(\boldsymbol{X}_2, \boldsymbol{U}_2\right)$, and $\left(\boldsymbol{X}_3, \boldsymbol{U}_3\right)$, corresponding to the ACM, GDM, and transition regions, respectively. Here, $\boldsymbol{X}_i$ and $\boldsymbol{U}_i$ denote the stacked state and input matrices for region $i$.

Training proceeds in three sequential phases. Each phase uses minibatch Stochastic Gradient Descent (SGD) to minimize a tailored subset of the loss terms introduced earlier.

\textbf{Phase 1 (ACM fitting):} Set $\lambda \equiv 0$ so that only the ACM contributes. For the ACM dataset $\left(\boldsymbol{X}_1, \boldsymbol{U}_1\right)$, forward predictions $\boldsymbol{\bar{x}}$ are computed using the ACM forward map (Eq.~\eqref{eq-X1}), with numerical integration (e.g., Runge–Kutta) producing the predicted trajectories $\boldsymbol{\bar{X}}_1$. The prediction error is evaluated via the loss $\mathcal{L}_{\text{model}}$. Minimizing this loss updates the ACM-related physical parameters, yielding the optimal estimate $\boldsymbol{\varphi}_{\text{ACM}}^{*}$. By design, when $\lambda = 0$, the loss induces zero gradient with respect to the GDM parameters, and therefore, the GDM parameters remain unchanged during Phase 1.

\textbf{Phase 2 (GDM fitting):} Set $\lambda \equiv 1$ so that only the GDM contributes. The same forward-prediction and integration pipeline is applied to the GDM dataset $\left(\boldsymbol{X}_2,\boldsymbol{U}_2 \right)$ to compute predictions $\boldsymbol{\tilde{X}}_2$. Minimizing the corresponding $\mathcal{L}_{\text{model}}$ yields the optimal GDM parameters $\boldsymbol{\varphi }_{\text{GDM}}^{*}$. Because $\lambda = 1$ in this phase, no gradients propagate to the ACM parameters, thereby preserving $\boldsymbol{\varphi }_{\text{ACM}}^{*}$. Upon completion of Phase~2, the full set of physical parameters $\boldsymbol{\varphi }^*$ is obtained.

\textbf{Phase 3 (Transition learning):} Finally, the physical parameters are held fixed $\boldsymbol{\varphi } = \boldsymbol{\varphi }^*$ and only the transition network parameters $\boldsymbol{\xi}$ are trained using the transition dataset $\left(\boldsymbol{X}_3,\boldsymbol{U}_3 \right)$. For each minibatch, mixed forward predictions $\boldsymbol{\hat{x}}$ are calculated, and the full hybrid loss $\mathcal{L}_{\text{total}}$ is evaluated. This loss includes the model prediction error, anchor-point constraints, monotonicity penalties, and smoothness regularization. Minimizing $\mathcal{L}_{\text{total}}$ via minibatch SGD yields the optimal transition network parameters $\boldsymbol{\xi^*}$. At the end of Phase 3, all components of the hybrid model have been identified, resulting in the complete parameter set $\left( \boldsymbol{\varphi^*},\boldsymbol{\xi^*} \right)$ that consistently captures aerodynamic behavior across all operating regimes.

Algorithm~\ref{algorithm-1} summarizes the proposed three-phase identification pipeline with parameter update steps.

\begin{algorithm}[tbp]
\caption{Three-Phase Parameter Identification for the Hybrid ACM--GDM Model}

\noindent \ \textbf{Input:} State matrices $\boldsymbol{X}_1$, $\boldsymbol{X}_2$, $\boldsymbol{X}_3$, for ACM, GDM, and transition regions respectively, Control input matrices $\boldsymbol{U}_1$, $\boldsymbol{U}_2$, $\boldsymbol{U}_3$, Dynamic model $\boldsymbol{h}$, First phase epochs $N_1$, Second phase epochs $N_2$, Third phase epochs $N_3$ \\
\noindent \ \textbf{Output:} Physical parameters $\boldsymbol{\varphi}^*$, transition network parameters $\boldsymbol{\xi}^*$

\begin{algorithmic}[1] 
\label{algorithm-1}

\STATE $k\leftarrow 0$
\STATE $\boldsymbol{\varphi} \leftarrow \boldsymbol{\varphi}_0$, $\boldsymbol{\xi} \leftarrow \boldsymbol{\xi}_0$
\WHILE{$k<N_1+N_2+N_3$}
    \STATE Generate the prediction using \eqref{eq-X1} if $k<N_1$, using \eqref{eq-X2} if $N_1\le k < N_1+N_2$, and using \eqref{eq-X3} otherwise
    
    \IF{$k<N_1+N_2$}
        \STATE Minimize the current data's loss \eqref{eq-loss} using mini-batch SGD, $\boldsymbol{\varphi} \leftarrow \boldsymbol{\varphi}_{k+1}$
    \ELSE
        \STATE Choose the anchor sets $\mathcal{A}_{\text{ACM}}$, $\mathcal{A}_{\text{GDM}}$ and the sampling grid $\mathcal{P}$
        \STATE Minimize the current adjoint loss \eqref{Loss_total} using mini-batch SGD, $\boldsymbol{\xi} \leftarrow \boldsymbol{\xi}_{k+1}$
    \ENDIF 
    \STATE $k\leftarrow k+1$
\ENDWHILE

\end{algorithmic}
\end{algorithm}

\section{Experiment}\label{section-Experiment}

\subsection{Experimental Setup}

\begin{figure*}[!t]
  \centering
  \subcaptionbox{Straight upward flight\label{subfig:straight-up}}{
    \includegraphics[width=0.195\linewidth]{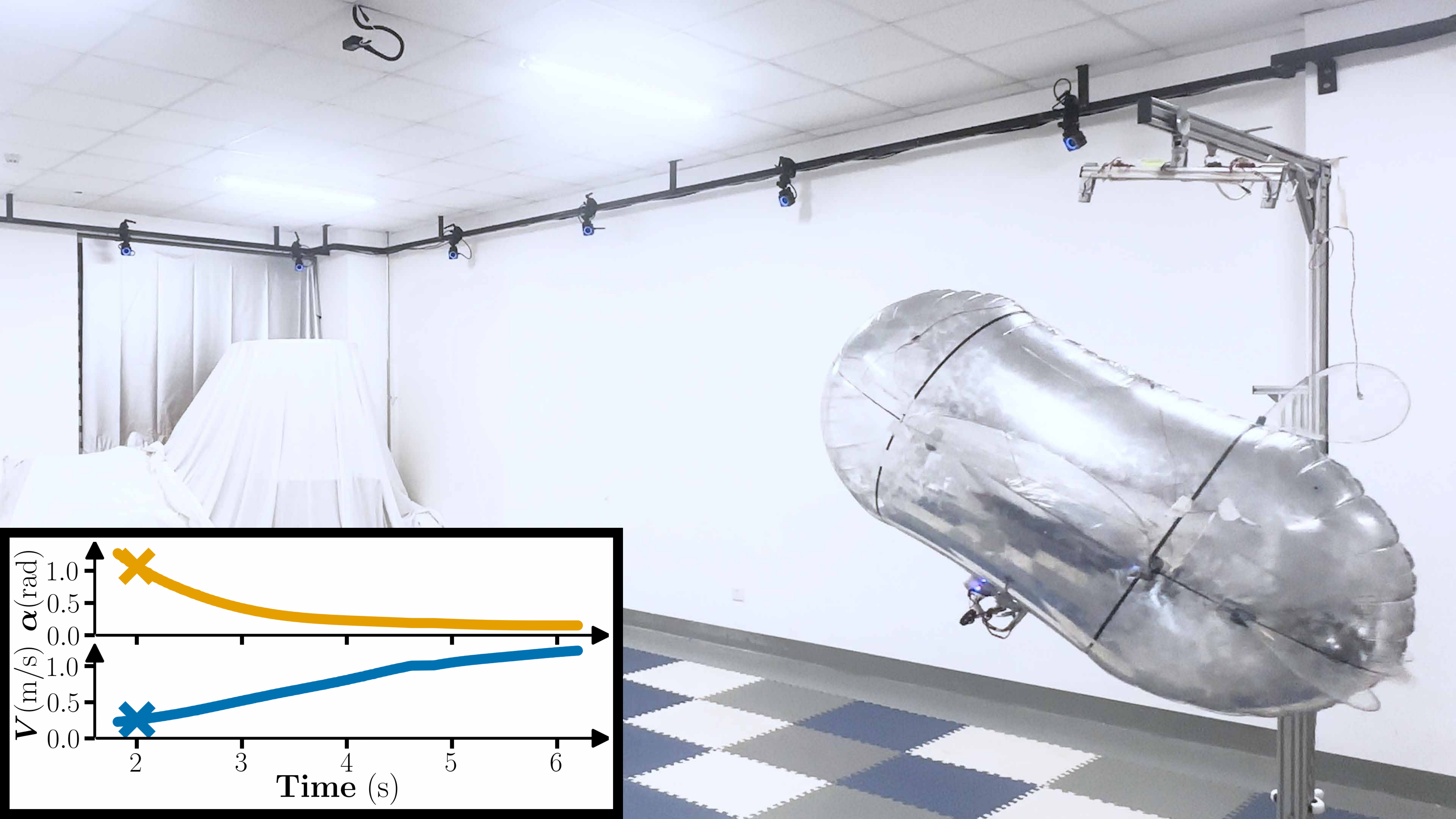}\hfill
    \includegraphics[width=0.195\linewidth]{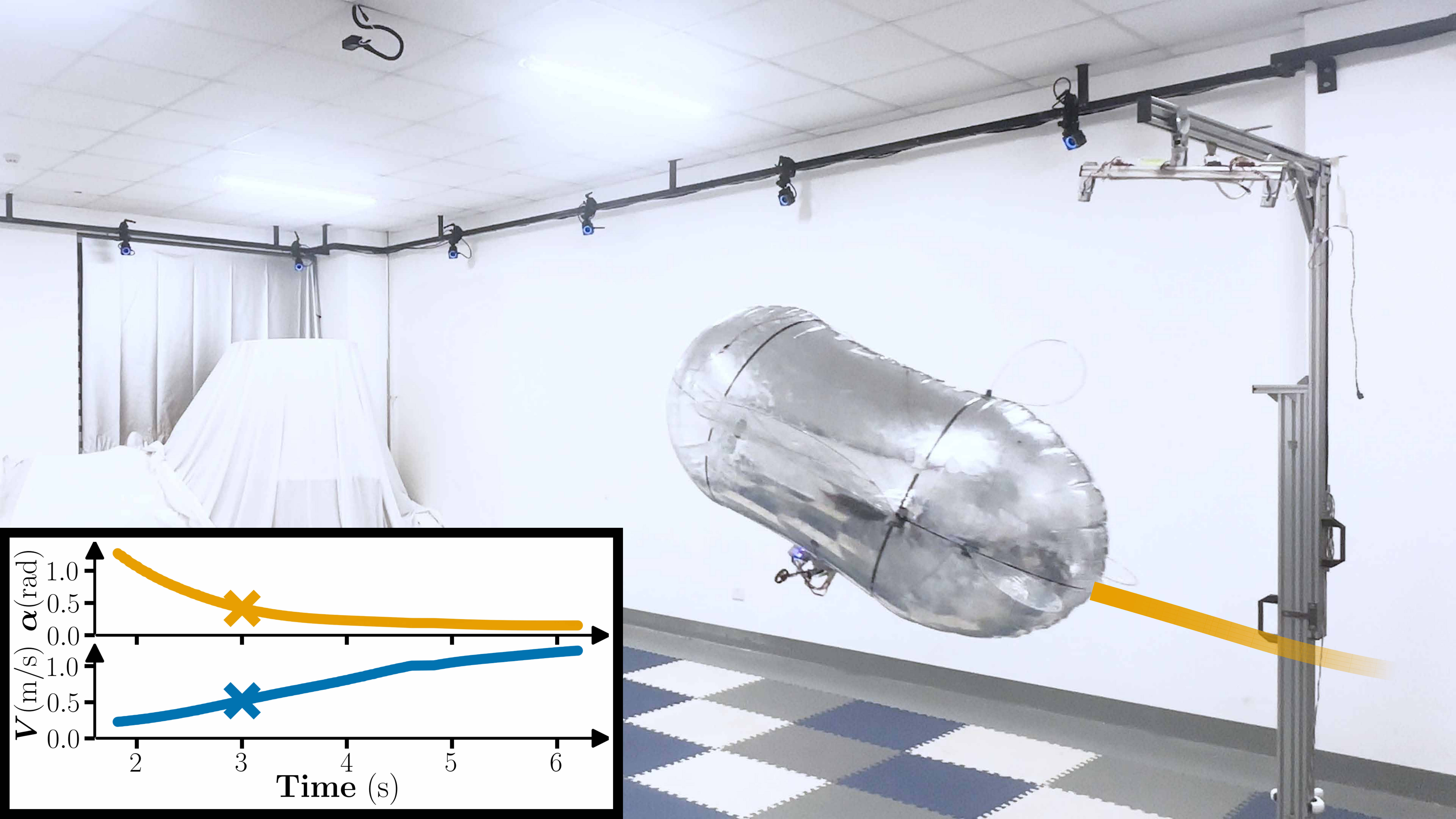}\hfill
    \includegraphics[width=0.195\linewidth]{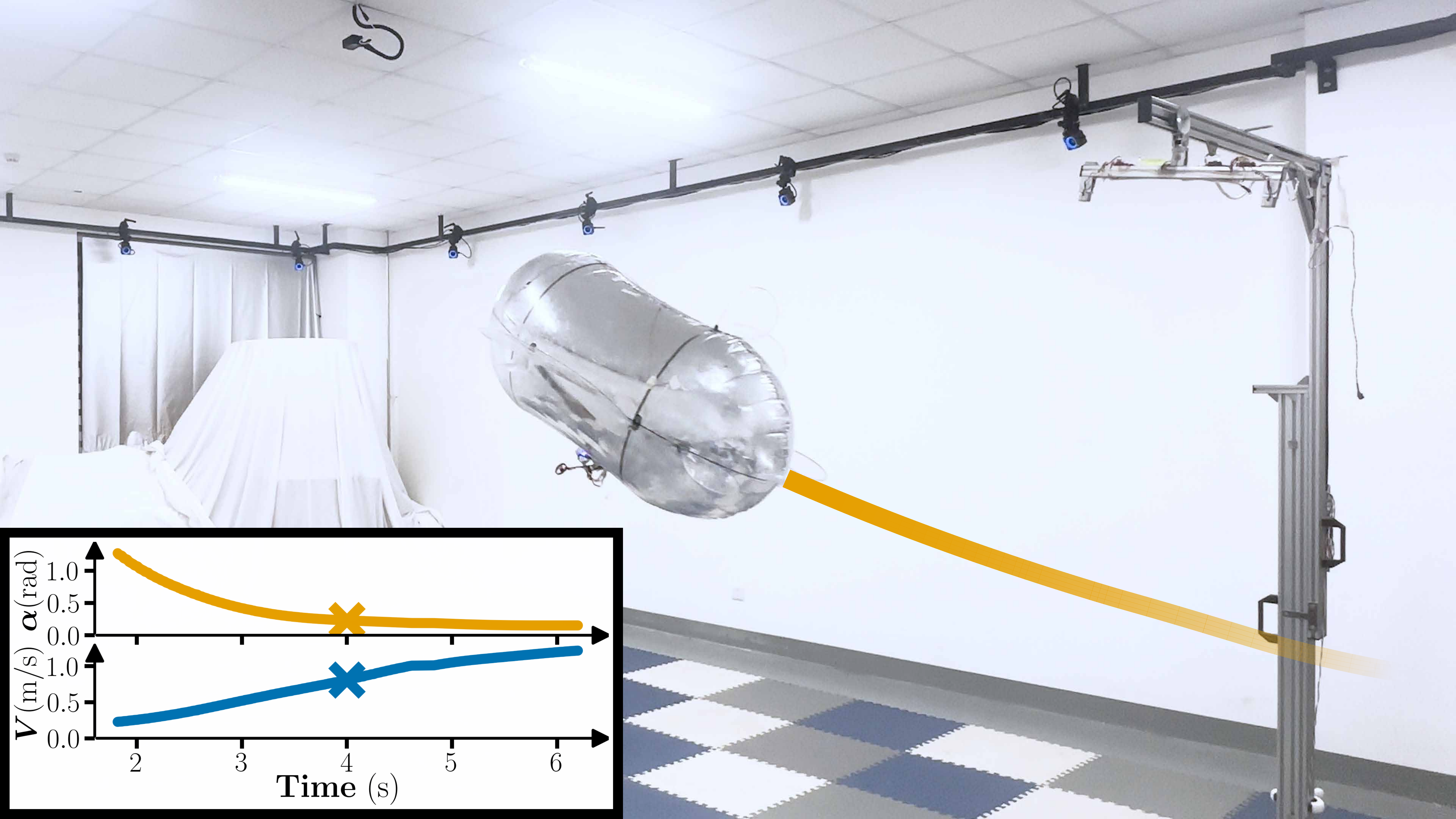}\hfill
    \includegraphics[width=0.195\linewidth]{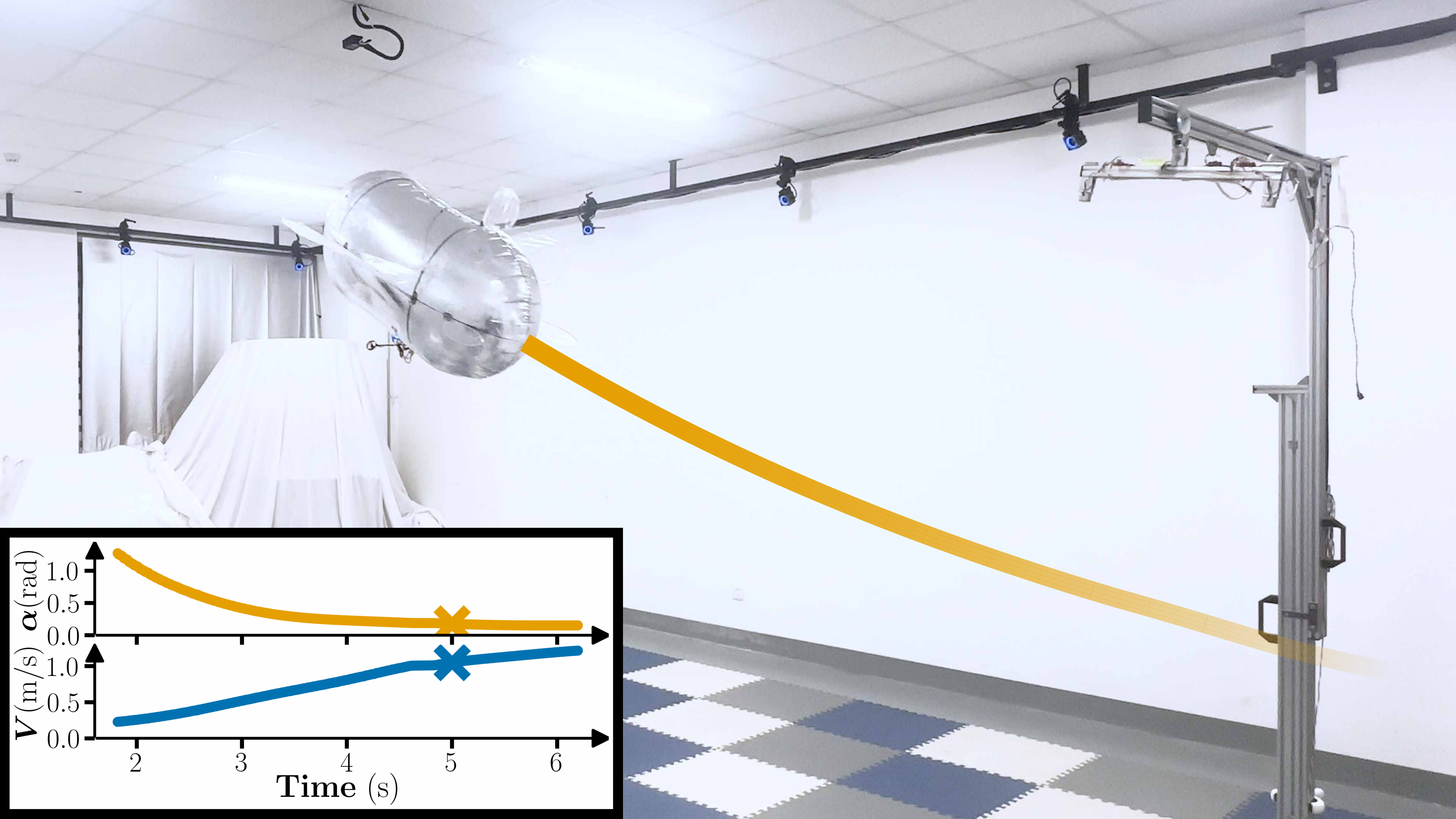}\hfill
    \includegraphics[width=0.195\linewidth]{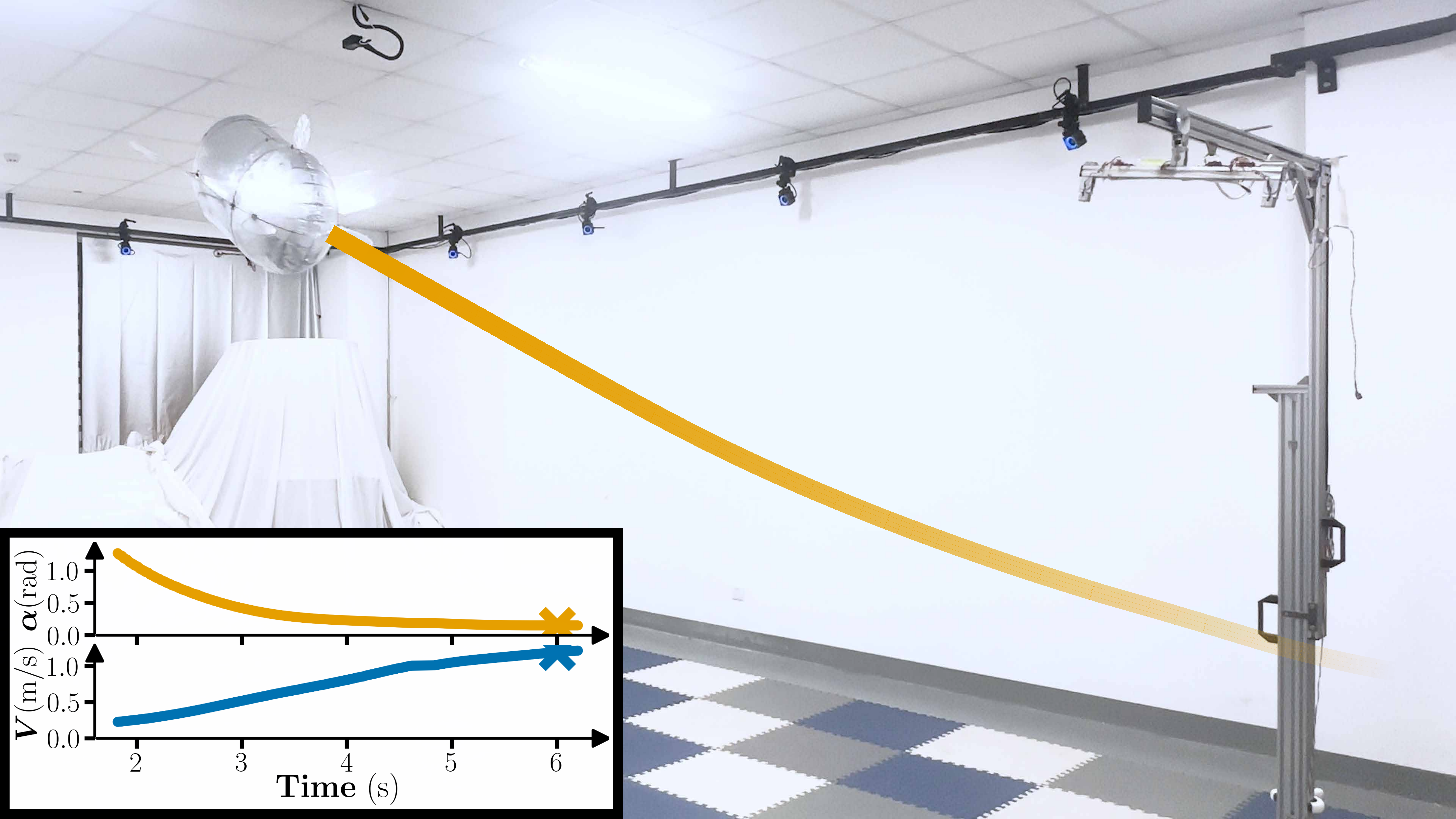}
  }\\[4pt]
  \subcaptionbox{Straight downward flight\label{subfig:straight-down}}{
    \includegraphics[width=0.195\linewidth]{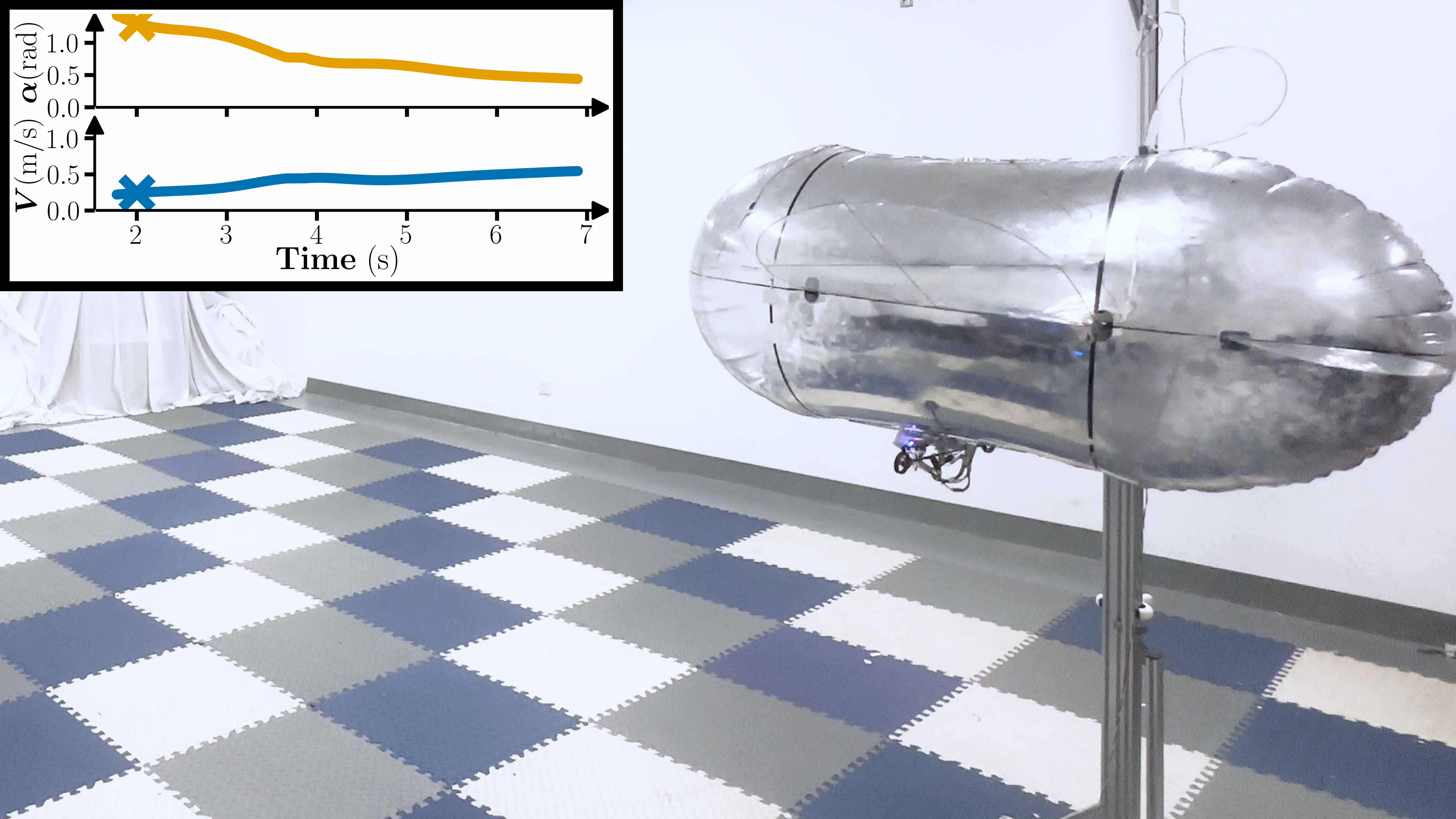}\hfill
    \includegraphics[width=0.195\linewidth]{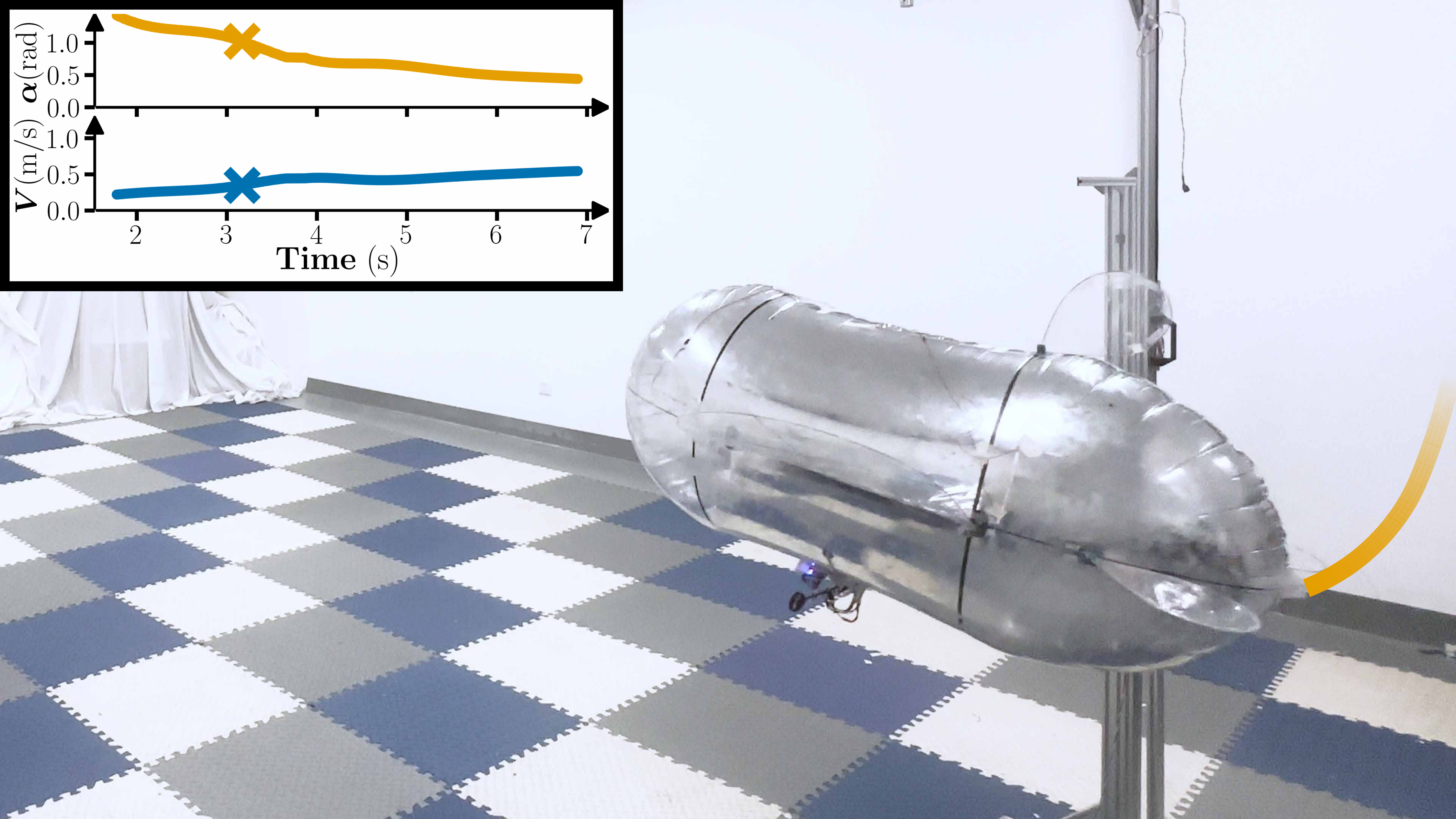}\hfill
    \includegraphics[width=0.195\linewidth]{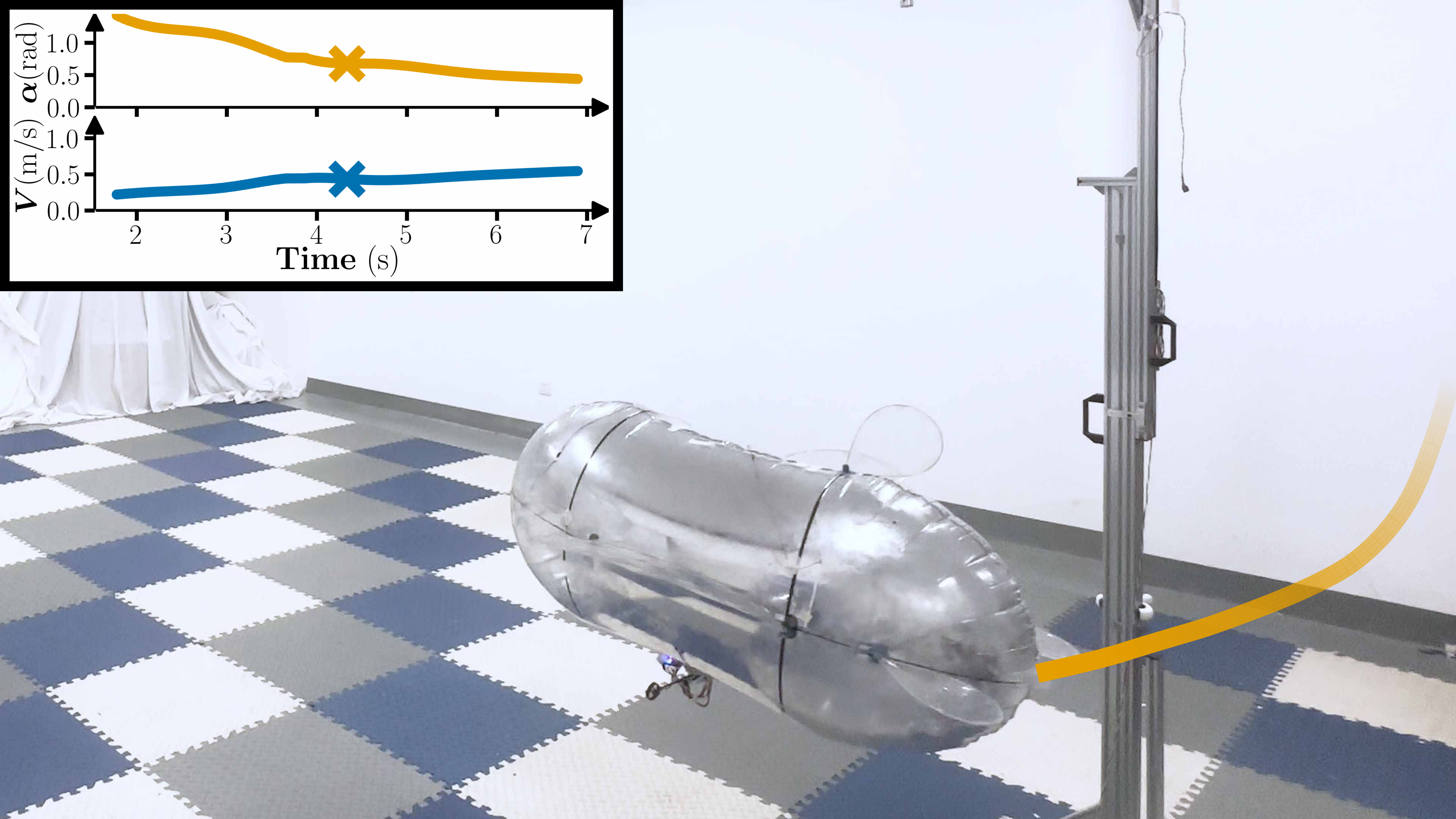}\hfill
    \includegraphics[width=0.195\linewidth]{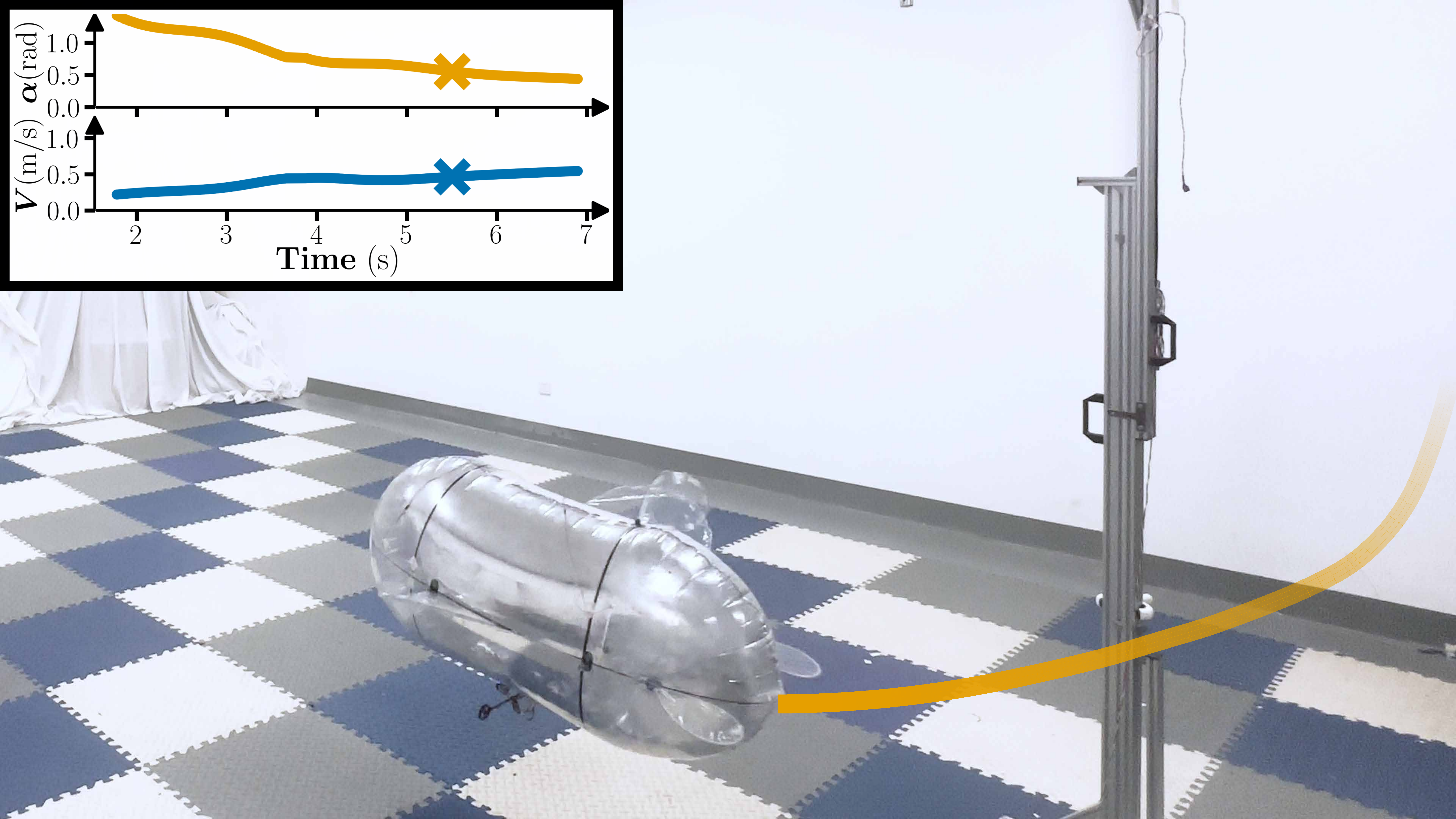}\hfill
    \includegraphics[width=0.195\linewidth]{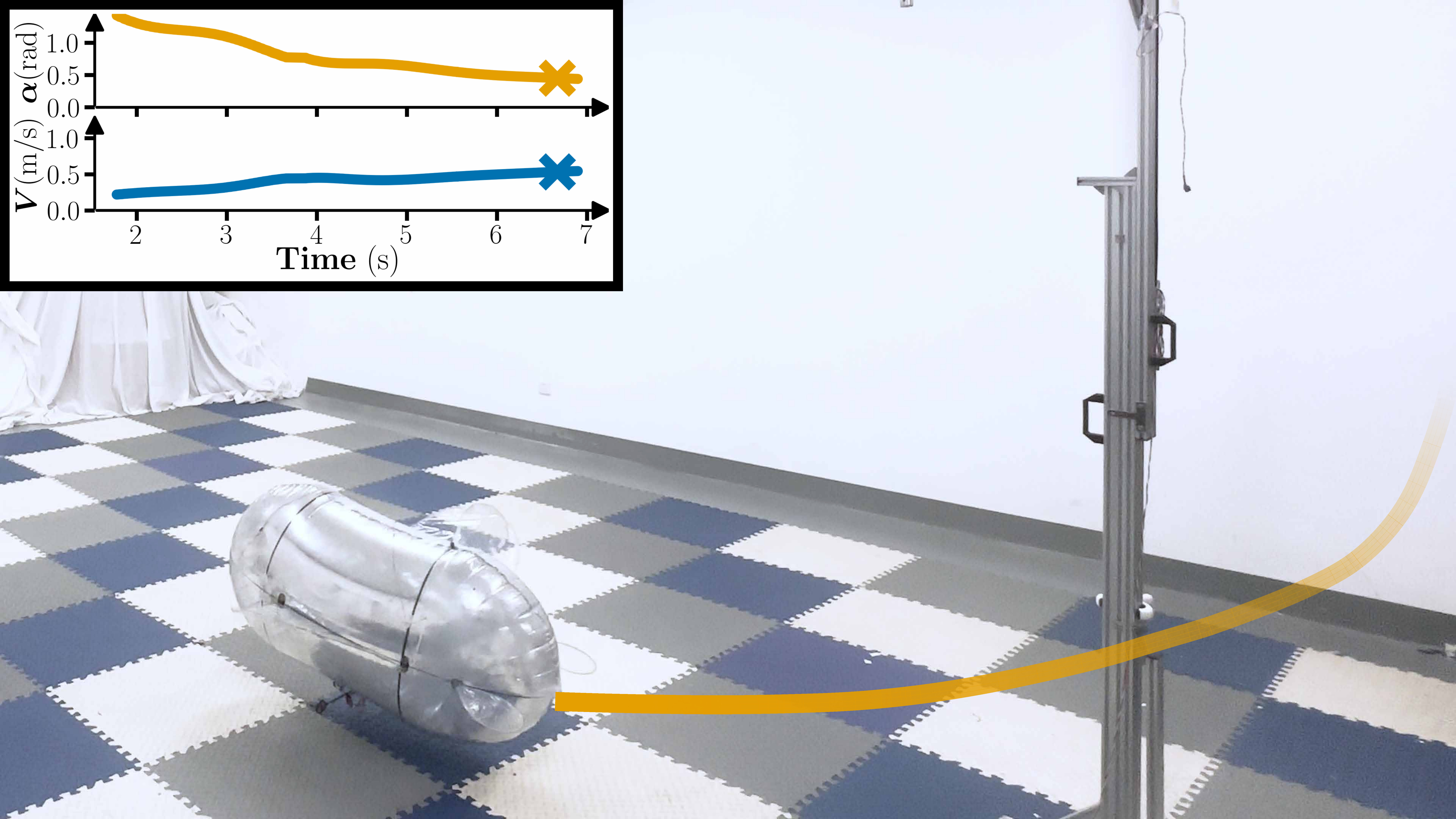}
  }\\[4pt]
  \subcaptionbox{Spiral upward flight\label{subfig:spiral-up}}{
    \includegraphics[width=0.195\linewidth]{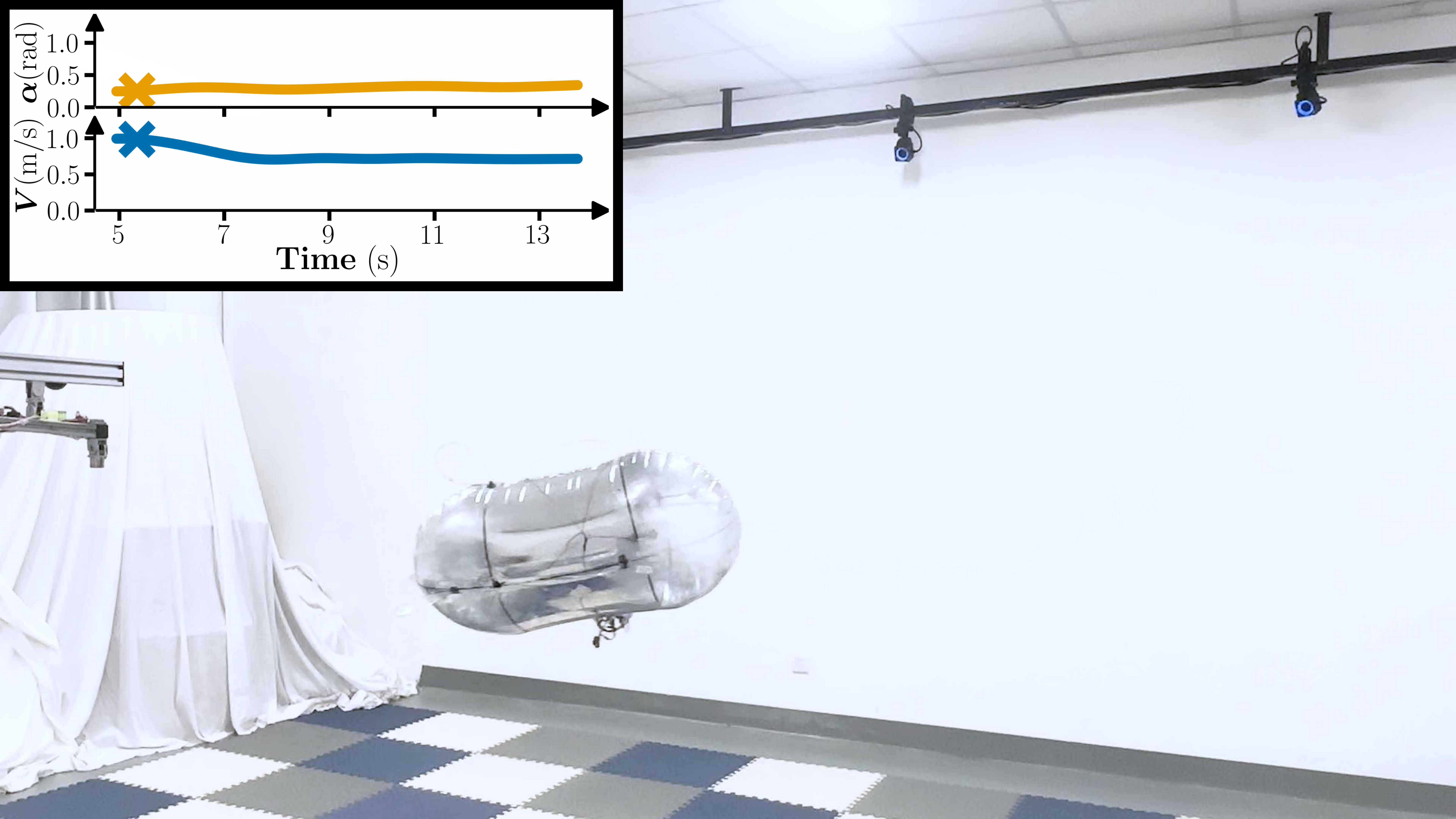}\hfill
    \includegraphics[width=0.195\linewidth]{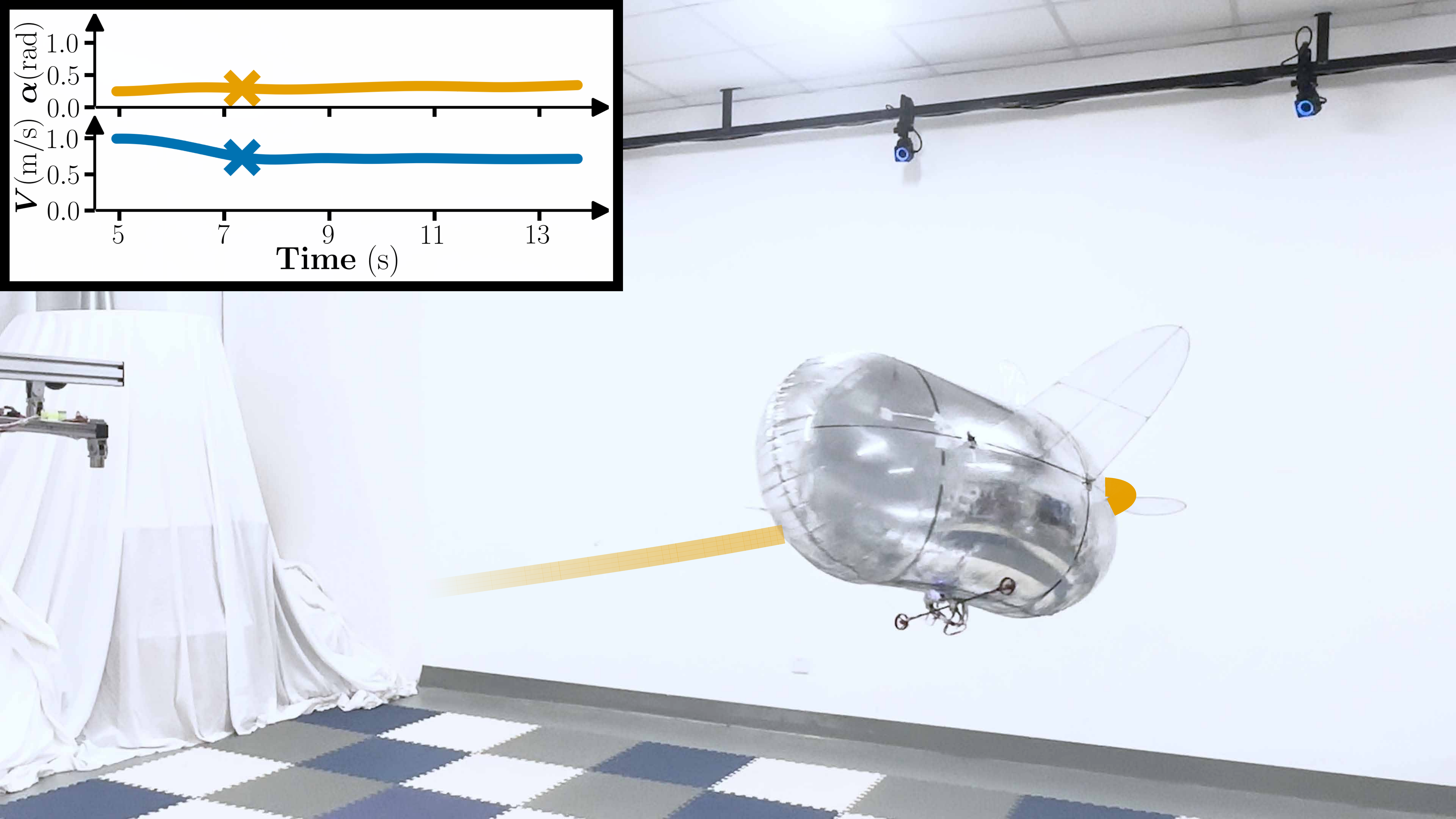}\hfill
    \includegraphics[width=0.195\linewidth]{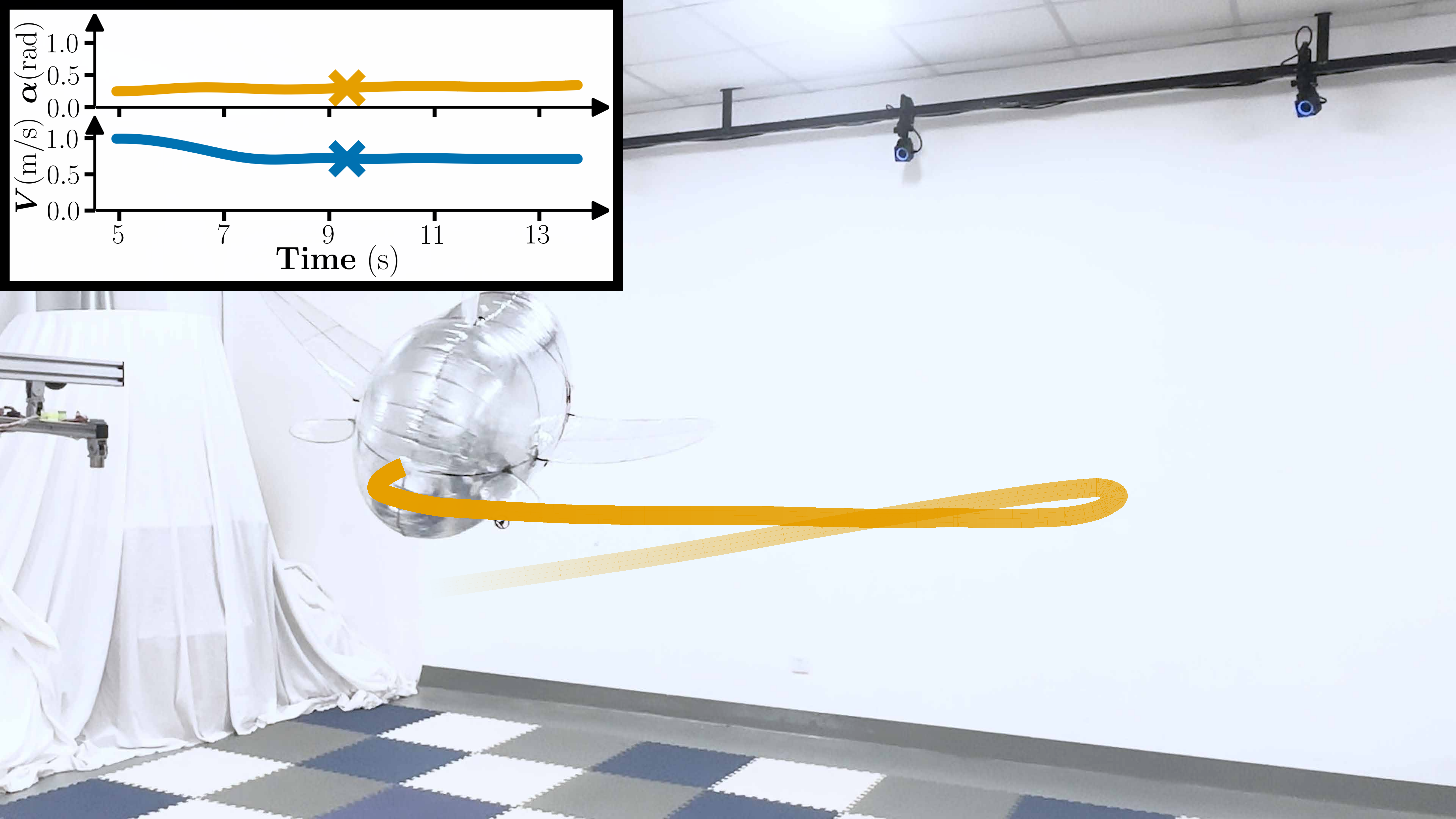}\hfill
    \includegraphics[width=0.195\linewidth]{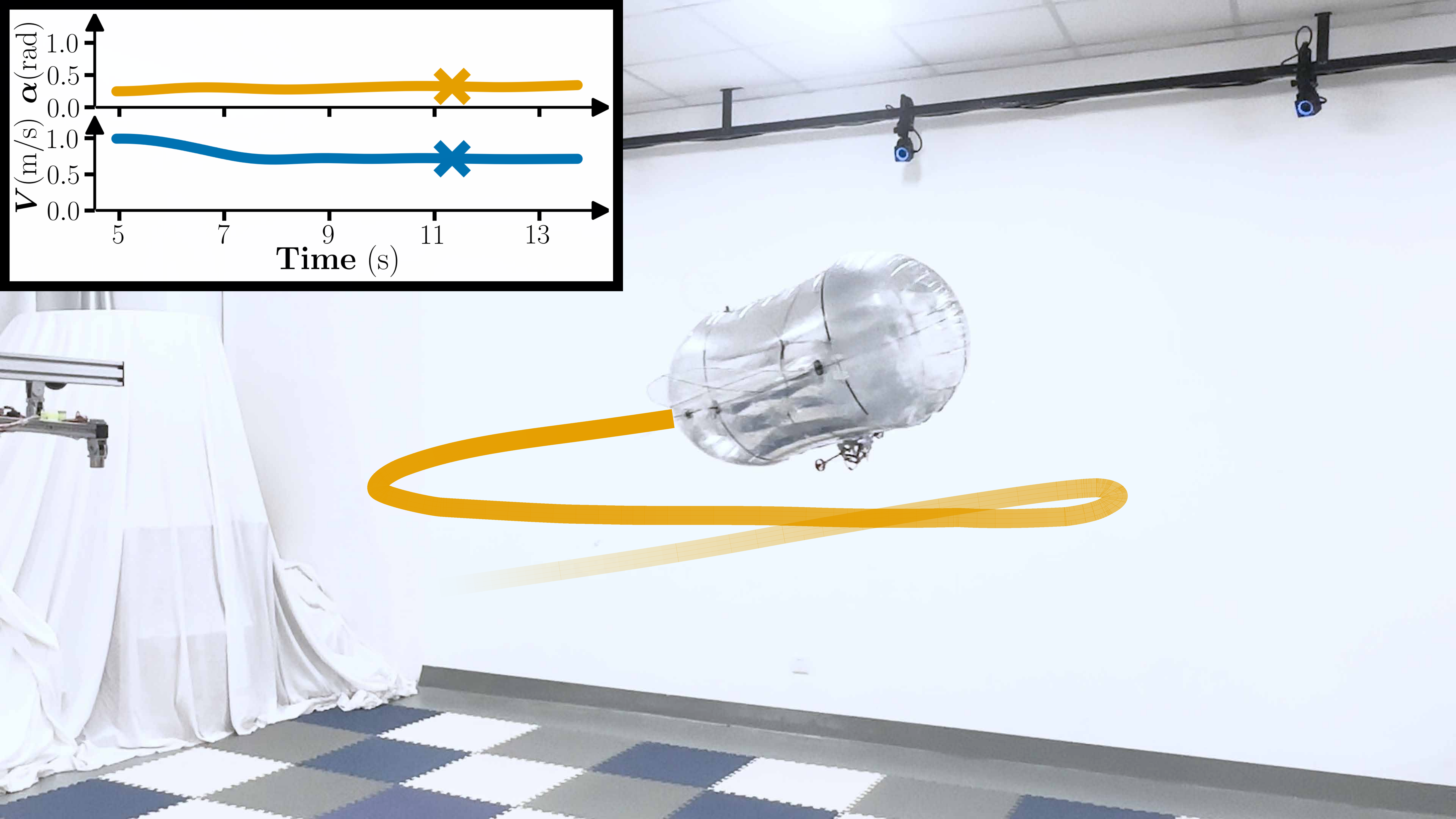}\hfill
    \includegraphics[width=0.195\linewidth]{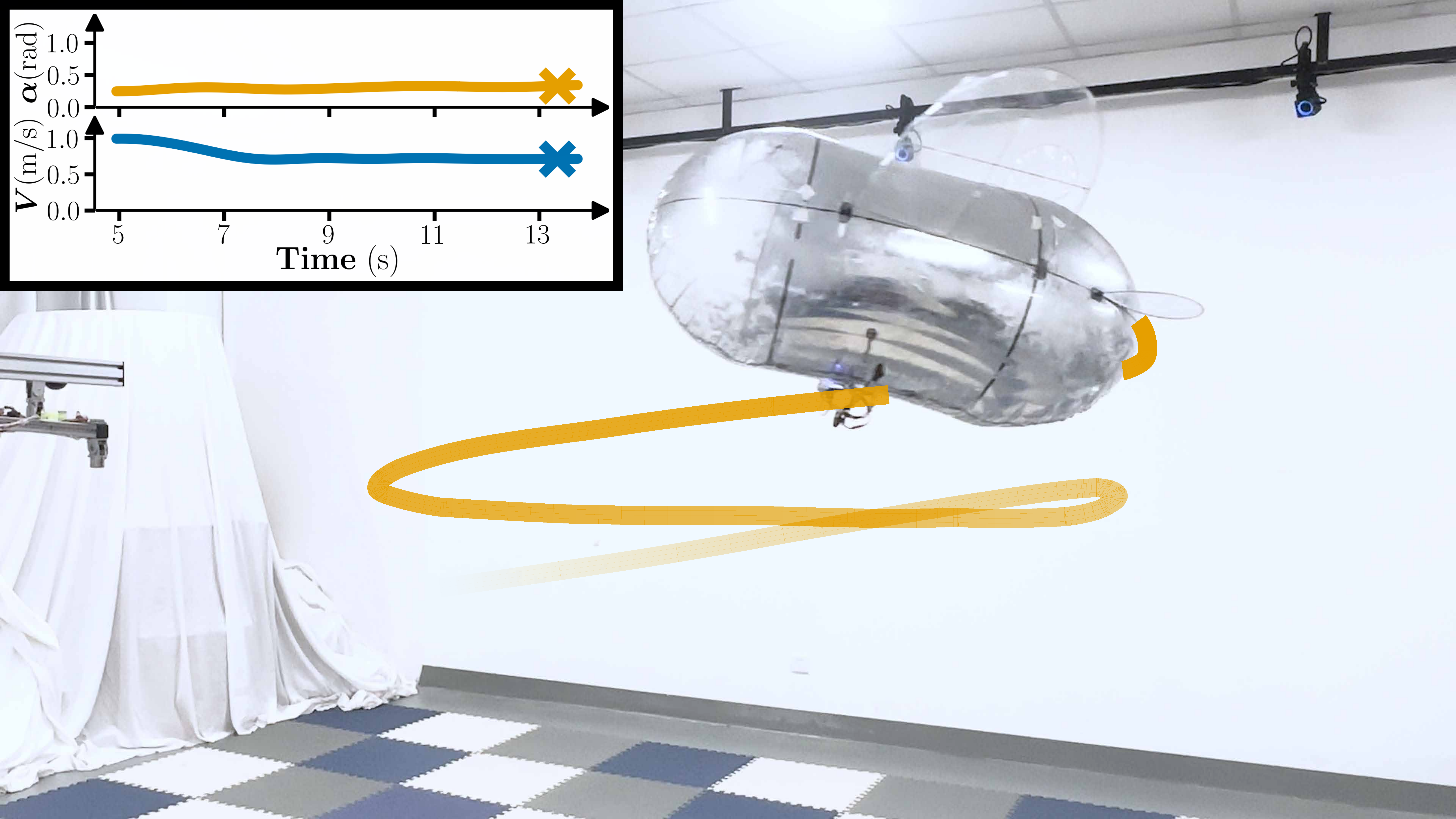}
  }\\[4pt]
  \subcaptionbox{Spiral downward flight\label{subfig:spiral-down}}{
    \includegraphics[width=0.195\linewidth]{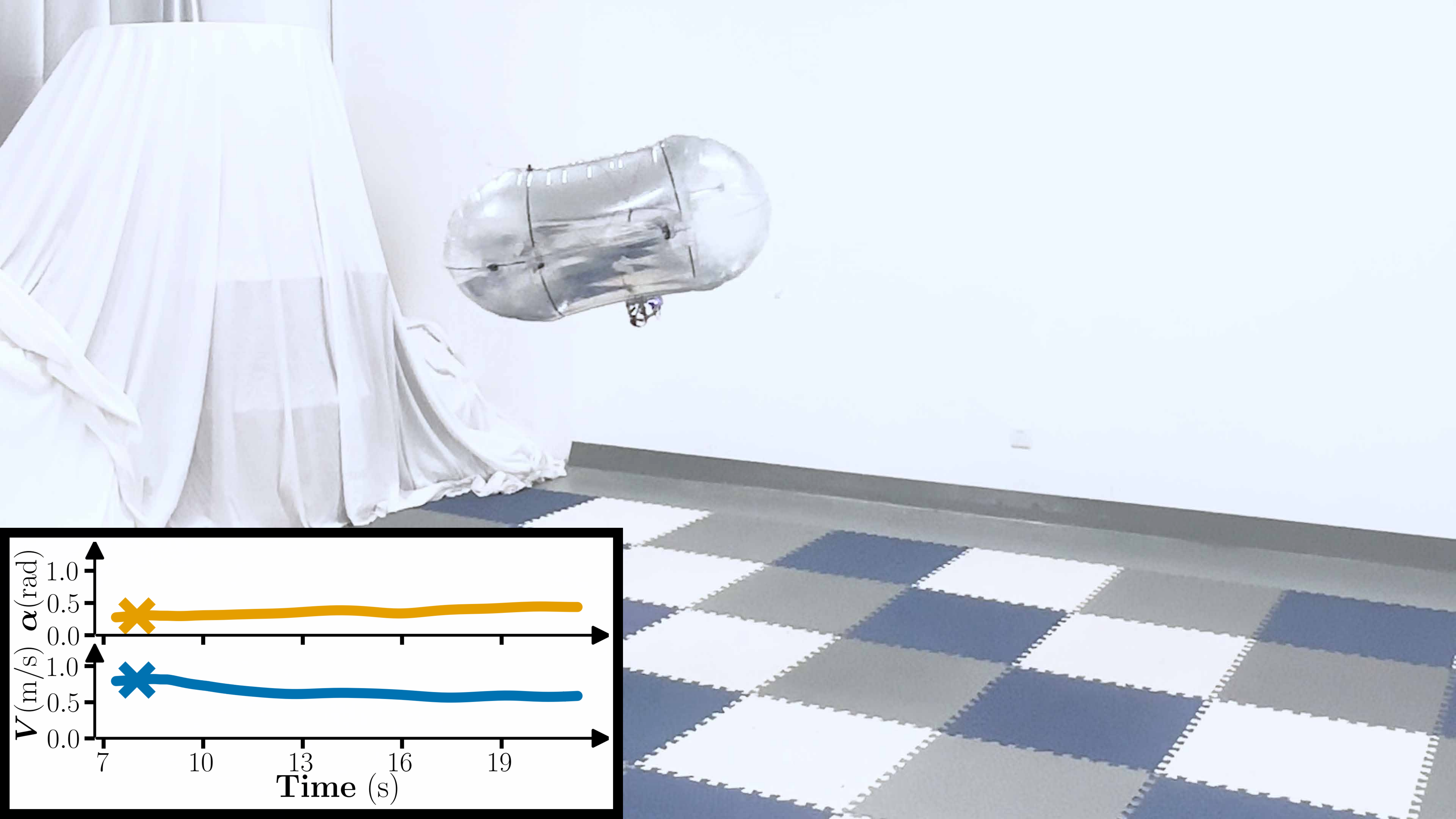}\hfill
    \includegraphics[width=0.195\linewidth]{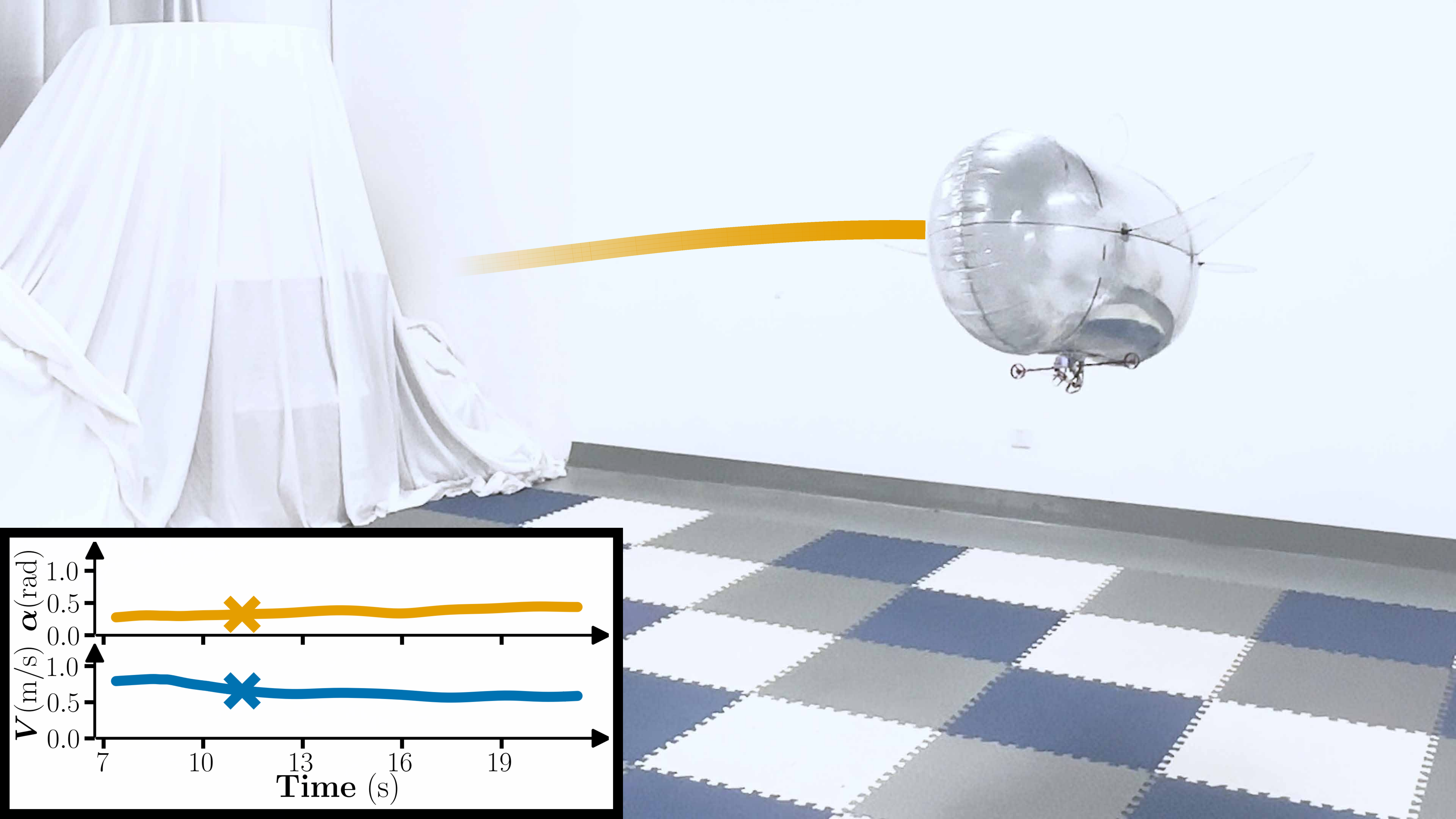}\hfill
    \includegraphics[width=0.195\linewidth]{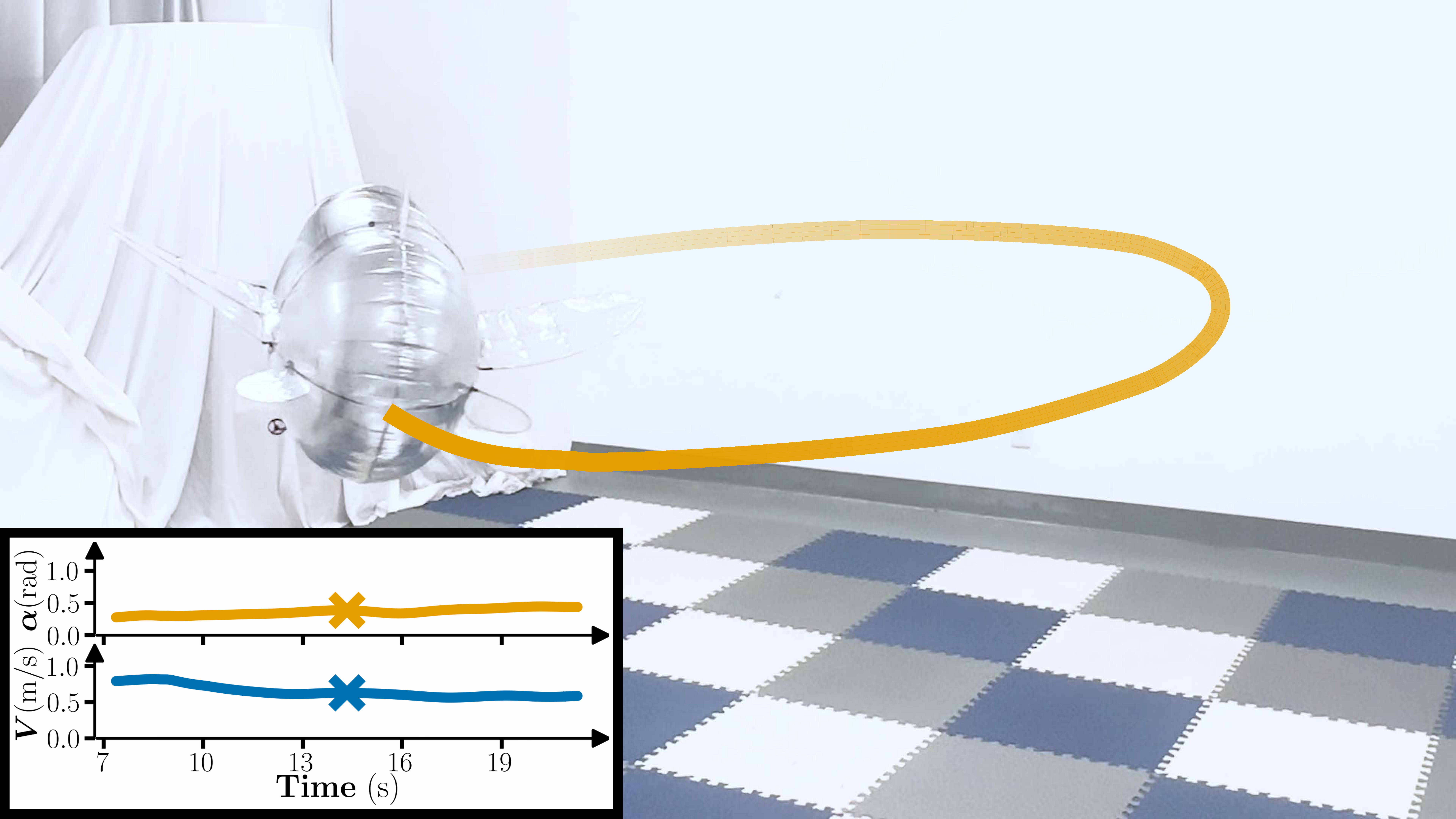}\hfill
    \includegraphics[width=0.195\linewidth]{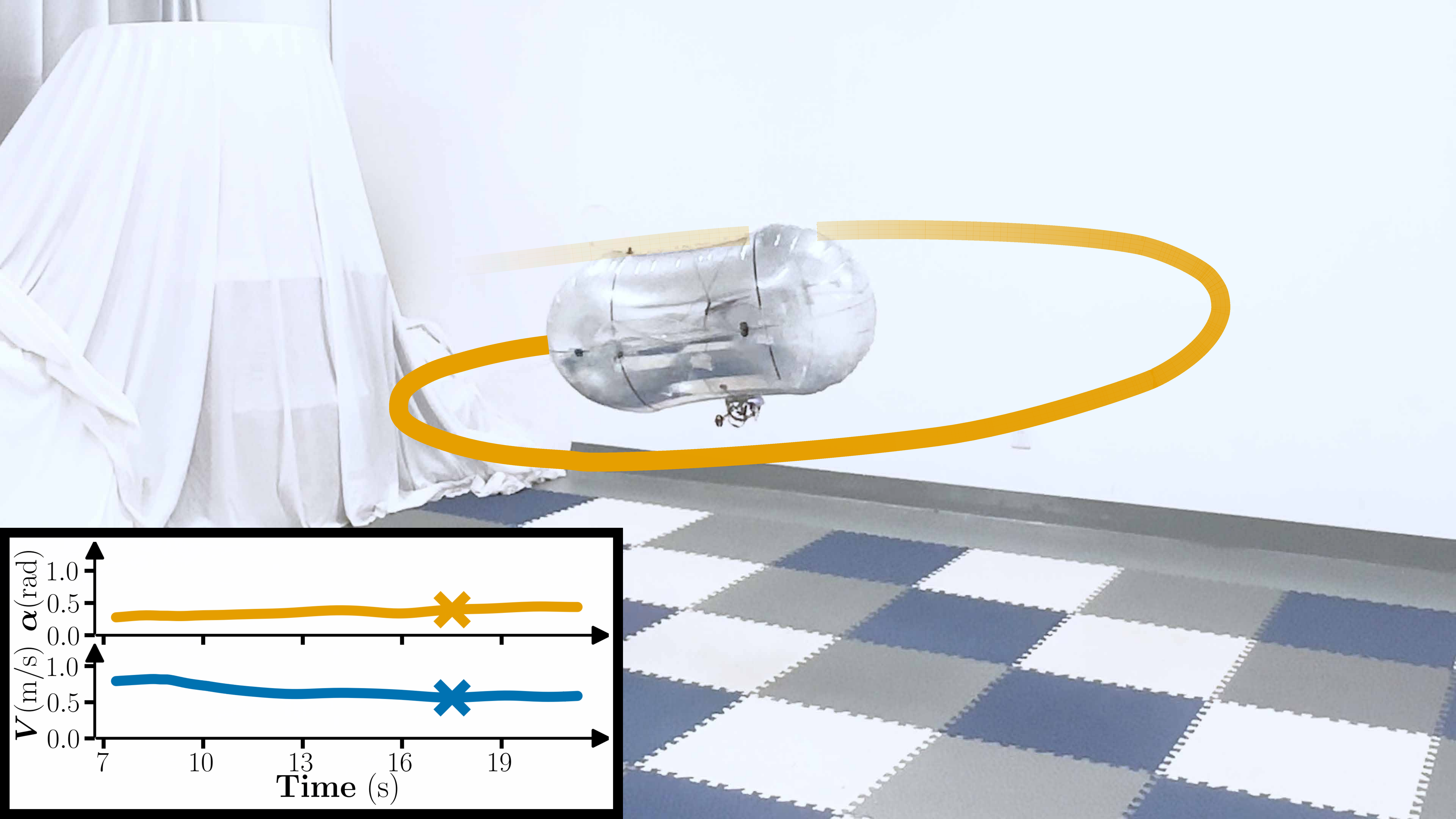}\hfill
    \includegraphics[width=0.195\linewidth]{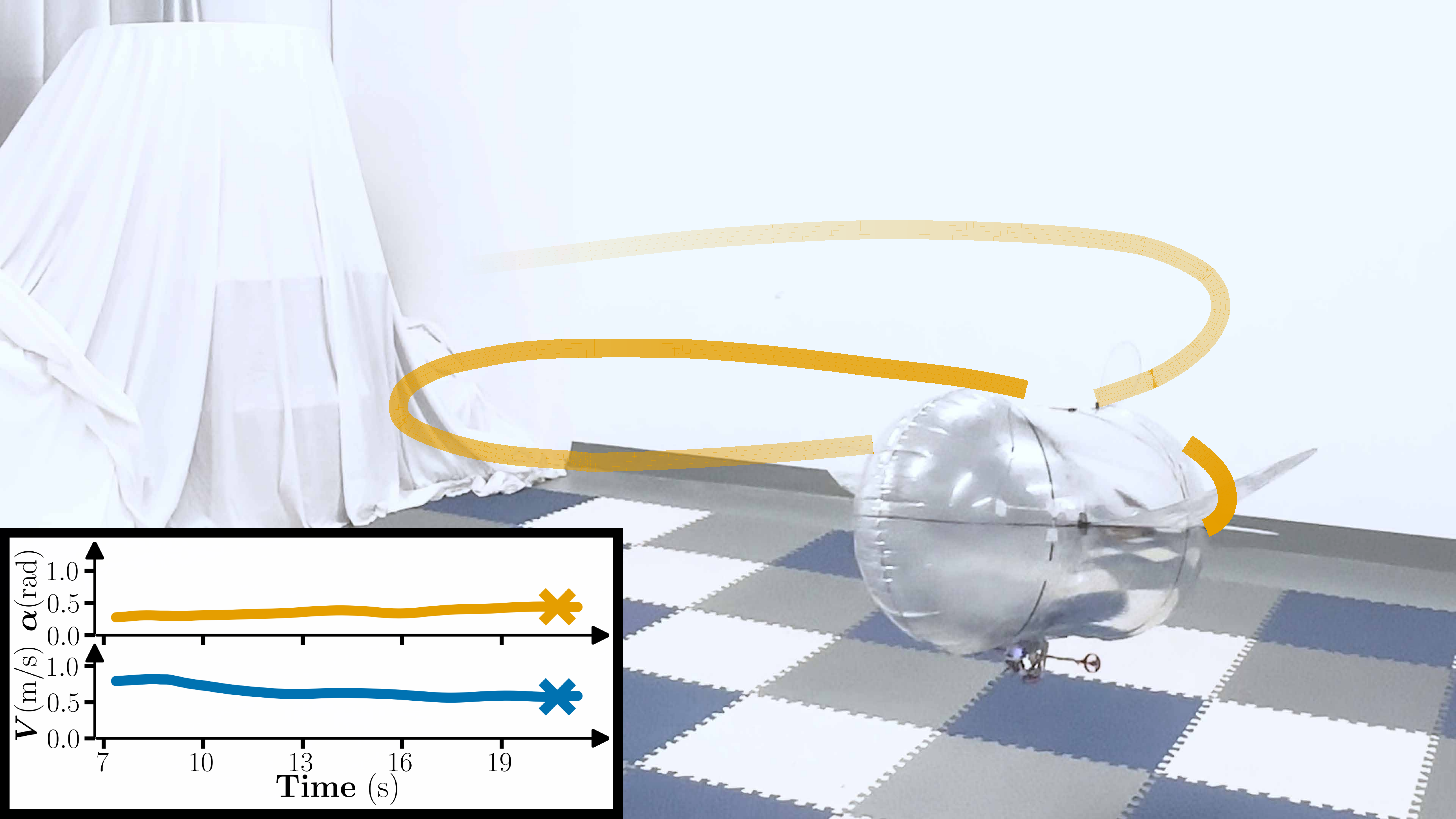}
  }
  \caption{Snapshots of RGBlimp flight experiments illustrating a variety of trajectories, angles of attack, and speeds. (a) Straight upward flight with input $\left (F_l,F_r,\Delta\bar{r}_x \right) = (4.59\,\text{gf},4.59\,\text{gf},-5\,\text{cm})$. (b) Straight downward flight with input $\left (F_l,F_r,\Delta\bar{r}_x \right) = (1.13\,\text{gf},1.13\,\text{gf},0\,\text{cm})$. (c) Spiral upward flight with input $\left (F_l,F_r,\Delta\bar{r}_x \right) = (7.57\,\text{gf},3.24\,\text{gf},0\,\text{cm})$. (d) Spiral downward flight with input $\left (F_l,F_r,\Delta\bar{r}_x \right) = (3.24\,\text{gf},2.06\,\text{gf},0\,\text{cm})$. Each snapshot includes an inset showing the time traces of $\alpha$ and $V$, along with their instantaneous values at the displayed frame.}
  \label{fig-snapshot}
\end{figure*}

Prior to flight testing, the blimp’s physical parameters that are independent of the aerodynamic coefficients were identified. A fixed quantity of helium was loaded into the envelope, setting the total mass $m$ and the buoyant force $B$. The envelope COM offset $\boldsymbol{r}_0$ was measured using a hanging method, and the initial position of the gondola moving mass, denoted by $\bar{\boldsymbol{r}}_0$, was defined as zero and recorded. The rigid-body inertia tensor of the envelope was obtained from the blimp CAD model.

All flight experiments were conducted in a motion-capture arena with dimensions 6\,m\,$\times$\,4\,m$\times$\,3\,m, equipped with 20 OptiTrack cameras. The motion-capture system operated at 60\,Hz and provided position estimates with approximately 0.76\,mm RMS accuracy. Twelve active markers were mounted on the envelope to provide continuous measurements of the blimp’s motion state.

Propulsion and onboard electronics are configured as follows. Each propeller is driven by a brushless DC motor (model SE0802-KV19000) and controlled by an electronic speed controller (ESC). The flight controller is built around a low-cost ESP8266 microcontroller running a rosserial\_arduino client. The controller exposes a ROS node over Wi-Fi for data logging and command streaming.

The experimental campaign spanned a grid of actuator and mass-trim configurations. We defined nine discrete ESC PWM levels (0--8) for each thruster, and characterized an approximately linear PWM-to-thrust mapping via static thrust measurements. The gondola moving mass translated along the body $O_bx_b$ axis with position
$\bar{\boldsymbol{r}} = \bar{\boldsymbol{r}}_0+\Delta \bar{r}_x \boldsymbol{i}_b$
where $\Delta \bar{r}_x$ took 11 equally spaced positions from $-5$\,cm to $+5$\,cm.

Flight tests included straight-line and spiral maneuvers. Straight-line trials used symmetric thruster settings, yielding 9 symmetric thrust configurations. Spiral trials used a fixed PWM difference of 1, 2, or 3 levels between the left and right thrusters, producing 21 differential-thrust configurations. To exploit approximate left--right symmetry, only configurations with the left thruster set higher than the right were tested. Each (thrust, moving-mass) pair was repeated 4 times, resulting in a total of \textbf{1,320} trajectories. The resulting dataset covers a broad range of angles of attack and speed magnitudes across straight climbs, straight descents, upward spirals, and downward spirals, as illustrated in Fig.~\ref{fig-snapshot}.

\subsection{Determining the Regime Switching Point}
\begin{figure*}[tbp]
\centerline{\includegraphics[width=1.0\linewidth]{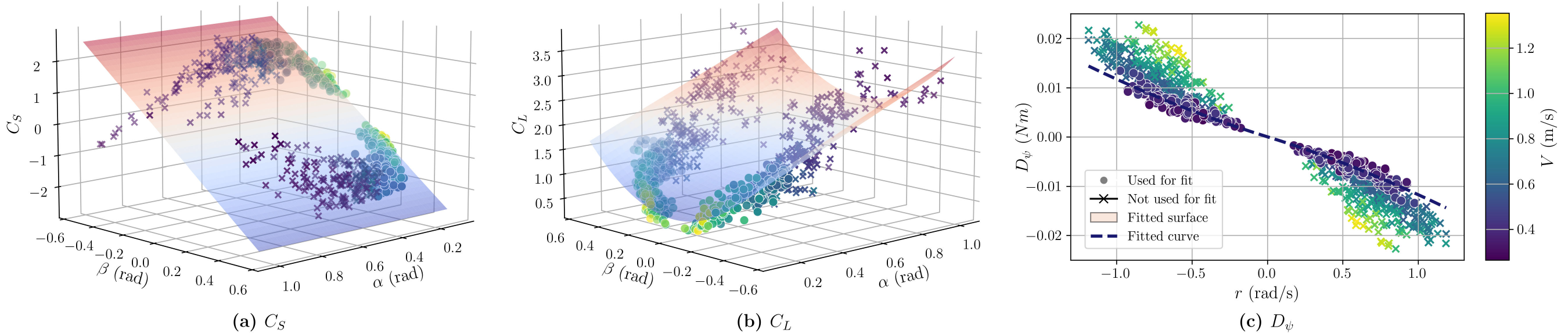}}
\caption{Steady-state flight data fitted using ordinary least squares to identify the model switching points $\alpha^*$ and $V^*$. (a) Fitted surface of the sideslip coefficient $C_S$ as a function of angle of attack $\alpha$ and sideslip angle $\beta$ in the ACM region. (b) Fitted surface of the lift coefficient $C_L$ as a function of $\alpha$ and $\beta$ in the ACM region. (c) Fitted relation of the yaw drag moment $D_{\psi}$ versus yaw rate $r$ in the GDM region.}
\label{fig-steady-fit}
\end{figure*}

Prior to training, we determined the switching thresholds $\alpha_1$, $\alpha_2$, $V_1$, and $V_2$ by analyzing the steady-state behavior of the hybrid dynamics in Eq.~\eqref{eq-hybrid}. Under the steady-state conditions, we computed the aerodynamic wrench as
\begin{equation}
\boldsymbol{F}_{\text{aero}}=\left( \boldsymbol{C}_{RB}\left( \boldsymbol{\nu } \right) +\boldsymbol{C}_A\left( \boldsymbol{\nu } \right) \right) \boldsymbol{\nu }
+\boldsymbol{g}\left( \boldsymbol{\eta } \right) -\boldsymbol{\tau }-\boldsymbol{\bar{F}}
\label{eq-faero-steady}
\end{equation}
Here, the state variables $\boldsymbol{\eta}$ and $\boldsymbol{\nu}$ were obtained directly from flight measurements. Immediately after motion onset, the velocity was approximately zero, enabling isolation of the acceleration-dependent fluid reaction for identifying the added-mass matrix. Using these estimates, we computed steady-state values of $\boldsymbol{F}_{\text{aero}}$ for all data points in the dataset.

Candidate functional forms of the ACM and the GDM were then used to represent $\boldsymbol{F}_{\text{aero}}$. We examined the ranges of $\alpha$ and $V$ over which each parametrization provided an adequate representation. From the experimental campaign, 398 steady-state points were extracted from spiral maneuvers. By exploiting left--right symmetry, this set was augmented to 796 steady-state samples. Both ACM and GDM parameterizations were fitted using ordinary least squares, and their fitting quality was evaluated across the $\alpha$ and $V$ domains.

The fitting results indicated that the ACM parameterization degraded when $\alpha$ was large or $V$ was small. Based on correlation and goodness-of-fit metrics, we determined critical switching points at $\alpha^*=0.40$\,rad and $V^*=0.45$\,m/s. By defining a transition band of $\pm 20\%$ around these values, we obtained $\alpha_1=0.32$\,rad, $\alpha_2=0.48$\,rad, $V_1=0.36$\,m/s, and $V_2=0.54$\,m/s. The identified angle threshold $\alpha^*=0.40$\,rad was physically reasonable, being close to typical stall-onset angles reported for small lifting surfaces \cite{high-alpha-2021, stall-2019}. For the velocity threshold $V^*=0.45$\,m/s, assuming $L=1$\,m, $\rho=1.225\,\text{kg/m}^3$, and $\mu =1.81\times 10^{-5}\,\text{Pa}\cdot \text{s}$, the corresponding Reynolds number was $\text{Re}=\rho VL/\mu \approx 3.05\times 10^4$, indicating that viscous effects were non-negligible \cite{2022-Re, fluids-2023-critical-Re}.

Figures \ref{fig-steady-fit}(a) and \ref{fig-steady-fit}(b) present representative coefficient fits in the ACM regime, while Fig.~\ref{fig-steady-fit}(c) shows a representative GDM-regime relation. In each case, samples within the designated region are well captured, whereas samples outside exhibit large deviations, supporting the selected thresholds and transition band.


\subsection{Model Training}
The dataset was partitioned into ACM, GDM, and transition subsets according to the identified switching thresholds. For the input grid $\left( F_l,F_r,\Delta\bar{r}_x \right)$, a total of 330 distinct configuration combinations were defined, each repeated four times. The resulting trajectories were randomly split into training and test sets at a 3:1 ratio while preserving the configuration distribution. Model training then followed the three-phase identification procedure described previously.

We employed a fourth-order Runge--Kutta (RK4) integrator for forward simulation and the Adam optimizer for parameter updates. The simulation time step matched the motion-capture rate of 60\,Hz. ACM and GDM parameters were initialized from the steady-state estimates obtained in the previous section. The transition network used three feedforward layers with hidden sizes of 32 and 16, ReLU activations, and a sigmoid output to constrain the mixing coefficient $\lambda$ to $\left[0,1\right]$. The learning rate was set to $10^{-3}$ for Phase 1 and Phase 2, and increased to $10^{-2}$ for Phase 3. Each phase was trained for $N_1=N_2=N_3=10$ epochs with minibatch updates.

\begin{figure}[tbp]
\centerline{\includegraphics[width=1.0\linewidth]{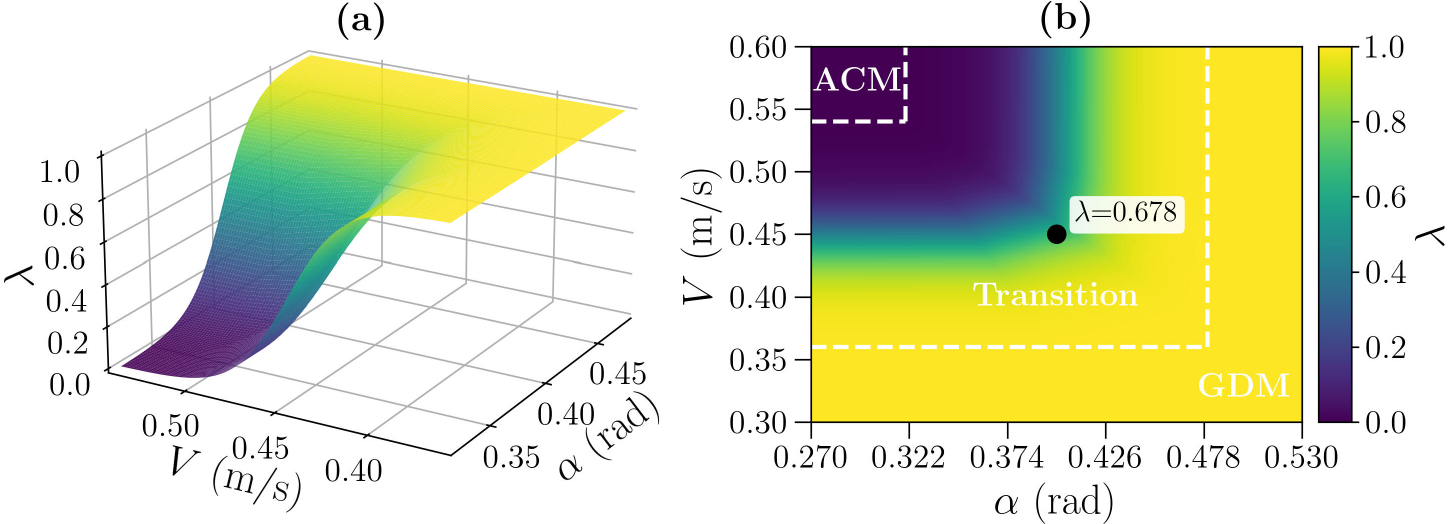}}
\caption{Learned mixing function $\lambda \left(\alpha,V \right)$ produced by the neural network. (a) Three-dimensional surface of $\lambda \left(\alpha,V \right)$. (b) Two-dimensional heatmap of $\lambda \left(\alpha,V \right)$. At the operating point $\left(\alpha^*,V^* \right) = \left( 0.40\,\text{rad}, 0.45\,\text{m/s}  \right)$ the network outputs $\lambda \approx 0.678$, indicating that the learned mixing is not strictly symmetric in $\alpha$ and $V$.}
\label{fig-lambda}
\end{figure}

Figure \ref{fig-lambda} illustrates the learned mixing function $\lambda\left(\alpha,V \right)$. The output remains close to 0 in the ACM region and close to 1 in the GDM region, while varying smoothly across the L-shaped transition band. The learned map is not perfectly symmetric, reflecting  asymmetries inherent in the experimental data rather than imposing artificial symmetry constraints.

\subsection{Comparison Models}
To validate and analyze the proposed hybrid model, we compare its performance against the following baselines.

\subsubsection{\textbf{ACM-only}} The Aerodynamic Coupling Model applied globally for all flight conditions.
\subsubsection{\textbf{GDM-only}} The Generalized Drag Model applied globally for all flight conditions.
\subsubsection{\textbf{ACM--GDM (hard)}} A piecewise hybrid model that applies ACM when $\alpha<\alpha^*$ and $V>V^*$, and otherwise switches to GDM with a binary mixing coefficient $\lambda \in \{0,1\}$.
\subsubsection{\textbf{ACM--GDM (sigmoid)}} A smooth hybrid model that blends ACM and GDM using a fixed, predefined sigmoid function of $\alpha$ and $V$.

In the following section, we compare these baselines with our proposed model \textbf{ACM--GDM (NN)} to quantify the benefits of aerodynamic hybridization and of a learned, physically regularized transition mechanism.

\section{Results and Analysis}\label{section-Results-and-Analysis}
\begin{figure*}[tbp]
\centerline{\includegraphics[width=1.0\linewidth]{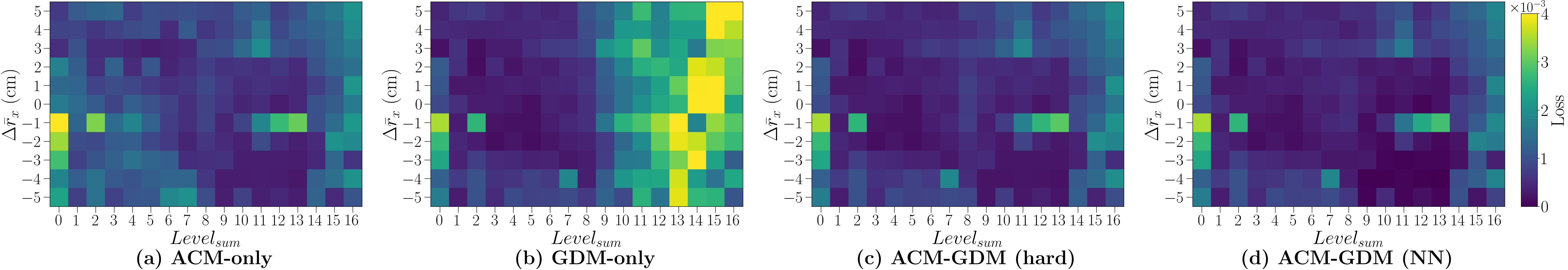}}
\caption{Heatmaps of the prediction loss $\mathcal{L}_{\text{model}}$ for ACM-only, GDM-only, ACM--GDM (hard) and ACM--GDM (NN) evaluated on the test set of 330 input configurations $\left( F_l,F_r,\Delta\bar{r}_x \right)$. The horizontal axis shows the combined thruster PWM level $Level_\text{sum}= Level_\text{l} +Level_\text{r}$ ranging from 0 to 16, serving as a proxy for total thrust. The vertical axis shows the moving-mass offset $\Delta\bar{r}_x$ with 11 setpoints from $-5$\,cm to $+5$\,cm. Each cell reports the average prediction loss for the corresponding input configuration across trials.}
\label{fig-loss-heatmap}
\end{figure*}

This section presents and analyzes the experimental results. Figure \ref{fig-loss-heatmap} summarizes predictive performance on the full test set. Four subplots report the forward-prediction loss $\mathcal{L}_\text{model}$ for the proposed ACM–GDM (NN) model and for three baselines: ACM-only, GDM-only, and ACM–GDM (hard). Each subplot represents a heatmap that aggregates results over all 330 thruster and moving-mass configurations $\left(F_l, F_r, \Delta \bar{r}_x\right)$ evaluated in the experiments. The horizontal axis corresponds to the combined thruster level $Level_\text{sum}=Level_l+Level_r$ (integer range 0–16), serving as a proxy for total thrust, while the vertical axis indicates the gondola moving-mass offset $\Delta\bar{r}_x$ (11 setpoints from $-5$\,cm to $+5$\,cm). The color of each cell represents the model’s forward-prediction loss for that corresponding configuration, thereby visualizing how prediction accuracy varies with thrust magnitude and trim condition.

For clarity, we evaluate model performance separately within the ACM region, the GDM region, and the transition region. In the following subsections, we provide detailed region-wise comparisons between the proposed hybrid model and each of the baseline methods, highlighting how prediction accuracy varies across the three operating regimes. 


\begin{table}[tb]
\renewcommand{\arraystretch}{1.25}
\caption{Average prediction RMSE ($\times 10^{-3}$) of each model in the ACM region, the GDM region, the transition region (mentioned in Section \ref{Section-Three-phase Training}), and the total across the entire flight envelope.}
\begin{center}
\begin{tabular}{@{}ccccc@{}}
\toprule
\multirow{2}{*}{\textbf{Model}}  & \multicolumn{4}{c}{\textbf{Region}} \\ \cmidrule(l){2-5}        & ACM  & GDM   & Transition    & Total  \\ \midrule
ACM-only         & 0.3946 & 1.0244 & 1.1429 & 0.7812 \\
GDM-only  & 2.0429 &0.4670 &1.0183 &1.2158 \\
ACM--GDM (hard)             & 0.3946 & 0.4670 & 0.7841 & 0.4892 \\
ACM--GDM (sigmoid)             & 0.3946 & 0.4670 & 0.5429 & 0.4494 \\ 
ACM--GDM (NN)        & 0.3946 & 0.4670 & 0.4495 & 0.4338 \\ 
\bottomrule
\end{tabular}
\label{tab-RMSE}
\end{center}
\end{table}

\subsection{ACM Region}
In the ACM region, the blimp operates at relatively high speeds and small angles of attack, where wing-generated aerodynamic lift–drag coupling is dominantly significant. As illustrated in Fig.~\ref{fig-loss-heatmap}, GDM-only performs poorly in this regime, particularly at large total thrust levels ($Level_\text{sum}>10$), when compared with models that employ ACM. Table~\ref{tab-RMSE} reports the average prediction loss by region. Specifically, in the ACM region, the average loss of GDM-only is approximately five times larger than that of the proposed ACM–GDM (NN). This result indicates that GDM fails to generalize under high-$V$, small-$\alpha$ conditions, and that explicitly modeling wing-related aerodynamic coupling is essential when such effects represent a dominant contribution to the dynamics.

\begin{figure}[tbp]
\centerline{\includegraphics[width=1.0\linewidth]{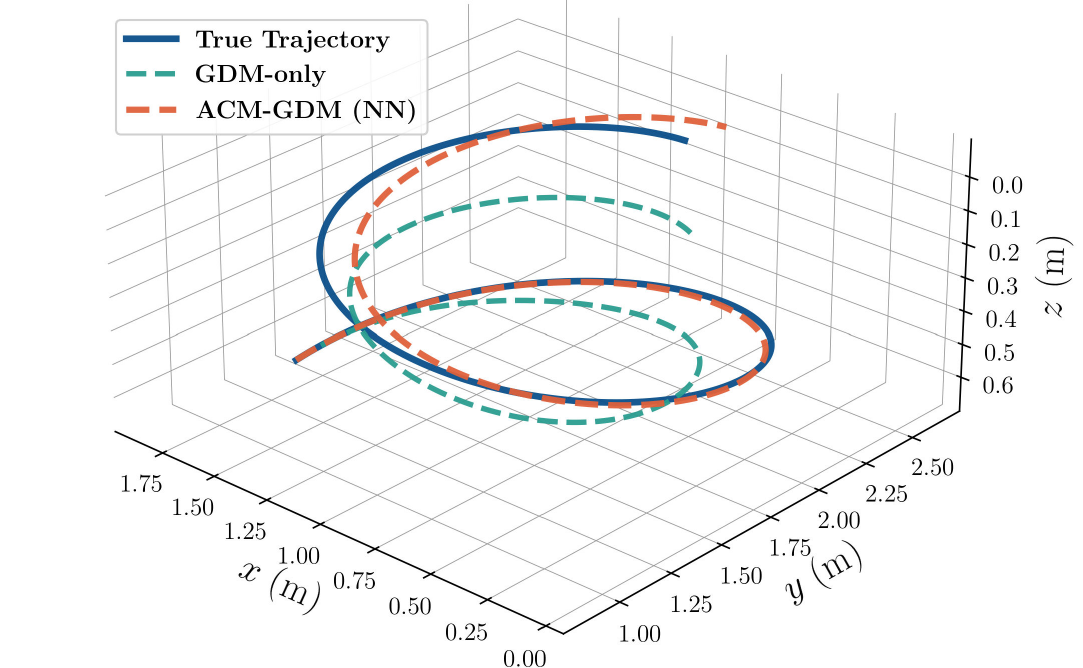}}
\caption{Spiral-climb trajectory for input $\left( F_l,F_r,\Delta\bar{r}_x \right)  = (6.05\,\text{gf},3.24\,\text{gf},0\,\text{cm})$ in the ACM region. Predicted paths from GDM-only and ACM–GDM (NN) are compared with the measured flight trajectory.}
\label{fig-trajectory-ACM}
\end{figure}

To illustrate this point more concretely, we consider a representative trajectory with input configuration $\left( F_l,F_r,\Delta\bar{r}_x \right) = (6.05\,\text{gf},3.24\,\text{gf},0\,\text{cm})$. Under this command, the blimp executes an upward spiral and eventually reaches a steady state characterized by high-$V$, small-$\alpha$ inside the ACM region. Figure \ref{fig-trajectory-ACM} compares the predicted trajectories of ACM–GDM (NN), which activates ACM in this regime, and the GDM-only baseline. Over the 8.5\,s maneuver, the proposed hybrid model closely tracks the measured trajectory, achieving a mean position error of 0.08\,m with only minor error accumulation at later times. By contrast, GDM-only deviates from the actual path at an early stage and yields a substantially larger mean error of 0.49\,m. 
Aerodynamically, the hybrid model predicts a steady-state speed of $V=0.68\,\text{m/s}$ and an angle of attack $\alpha=0.19\,\text{rad}$, resulting in an estimated aerodynamic lift of 8.19\,gf. Given the blimp’s net weight of approximately 6.25\,gf, this lift is clearly significant and provides the primary mechanism for the observed climb motion. Consistent with this interpretation, GDM-only predicts a weak ascent with a final vertical speed of 0.036\,m/s, whereas ACM–GDM (NN) predicts a sustained vertical ascent at 0.075\,m/s, in closer agreement with the experimental measurements. These results further confirm the necessity of incorporating ACM effects in the high-$V$, small-$\alpha$ regime.

\subsection{GDM Region}
In the GDM region, the blimp operates at low speeds or at very large angles of attack, where wing-generated lift becomes negligible. Bluff-body-like forces dominate at large angles, and viscous drag is significant at low speeds. The aerodynamic behavior in this regime is well described by drag-dominated, envelope-style models typical of conventional indoor blimps. Accordingly, the GDM formulation substantially improves model accuracy. As illustrated in Fig.~\ref{fig-loss-heatmap}, for $Level_\text{sum}<8$ models using GDM outperform ACM-only. Table~\ref{tab-RMSE} further presents that the average loss of the proposed ACM–GDM hybrid in the GDM region is approximately half of the ACM-only loss. These results confirm the necessity of incorporating GDM in low-$V$ / high-$\alpha$ conditions.

\begin{figure}[tbp]
\centerline{\includegraphics[width=1.0\linewidth]{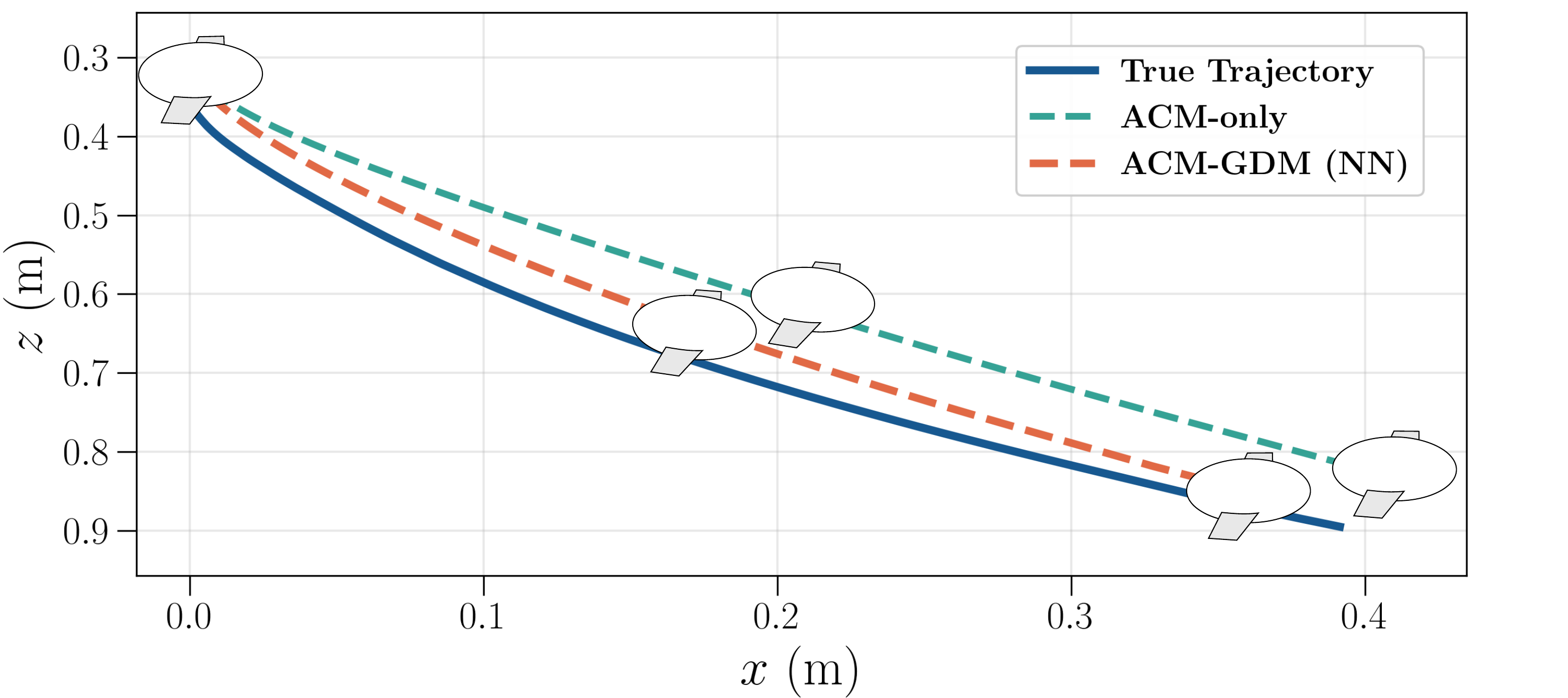}}
\caption{Straight downward trajectory on the $xz$ plane for  input $\left( F_l,F_r,\Delta\bar{r}_x \right)  = (1.13\,\text{gf},1.13\,\text{gf},0\,\text{cm})$ in the GDM region. Predicted paths from ACM-only and ACM–GDM (NN) are compared with the measured flight trajectory.}
\label{fig-trajectory-GDM}
\end{figure}

Consider the configuration $\left( F_l,F_r,\Delta\bar{r}_x \right) = (1.13\,\text{gf},1.13\,\text{gf},0\,\text{cm})$ as a concrete example. Under this input, the blimp executes an almost-vertical descent with a minimal forward component. The flight is characterized by a very large angle of attack and low speed, resulting in bluff-body-like flow and minimal effective lift from the wings. Figure~\ref{fig-trajectory-GDM} presents the 3\,s trajectory projected onto the $xz$ plane. In this case, the ACM--GDM hybrid, which uses GDM in this regime, attains a trajectory error of 0.033\,m, closely matching the measured path, while ACM-only yields a larger error of 0.048\,m. The discrepancy is due to that the ACM model predicts an aerodynamic lift of 3.68\,gf at $\alpha>0.8$\,rad. That lift, when transformed from the velocity frame to the inertial frame, produces a horizontal component of approximately 1.8\,gf, inducing an erroneous forward drift. By employing GDM in this regime, the hybrid model avoids this spurious lift prediction and more accurately reproduces the observed motion.

\subsection{Transition Region}
Focusing on the transition region, Fig.~\ref{fig-loss-heatmap} illustrates that configurations with intermediate total thrust, approximately $8\le Level_\text{sum}\le 10$, often place the blimp within the transition band. In this region, ACM–GDM (NN) achieves noticeably lower prediction loss compared with ACM–GDM (hard). Table~\ref{tab-RMSE} quantifies this improvement. Specifically, the prediction loss of ACM–GDM (NN) in the transition region is approximately 42.7\% lower than that of ACM–GDM (hard) and about 20.8\% lower than that of ACM–GDM (sigmoid). These results indicate that a smooth, learned blending mechanism outperforms both a binary switch and a fixed analytic interpolant, while satisfying the physical regularizers of anchor points, monotonicity, and smoothness.

\begin{figure}[tbp]
\centerline{\includegraphics[width=1.0\linewidth]{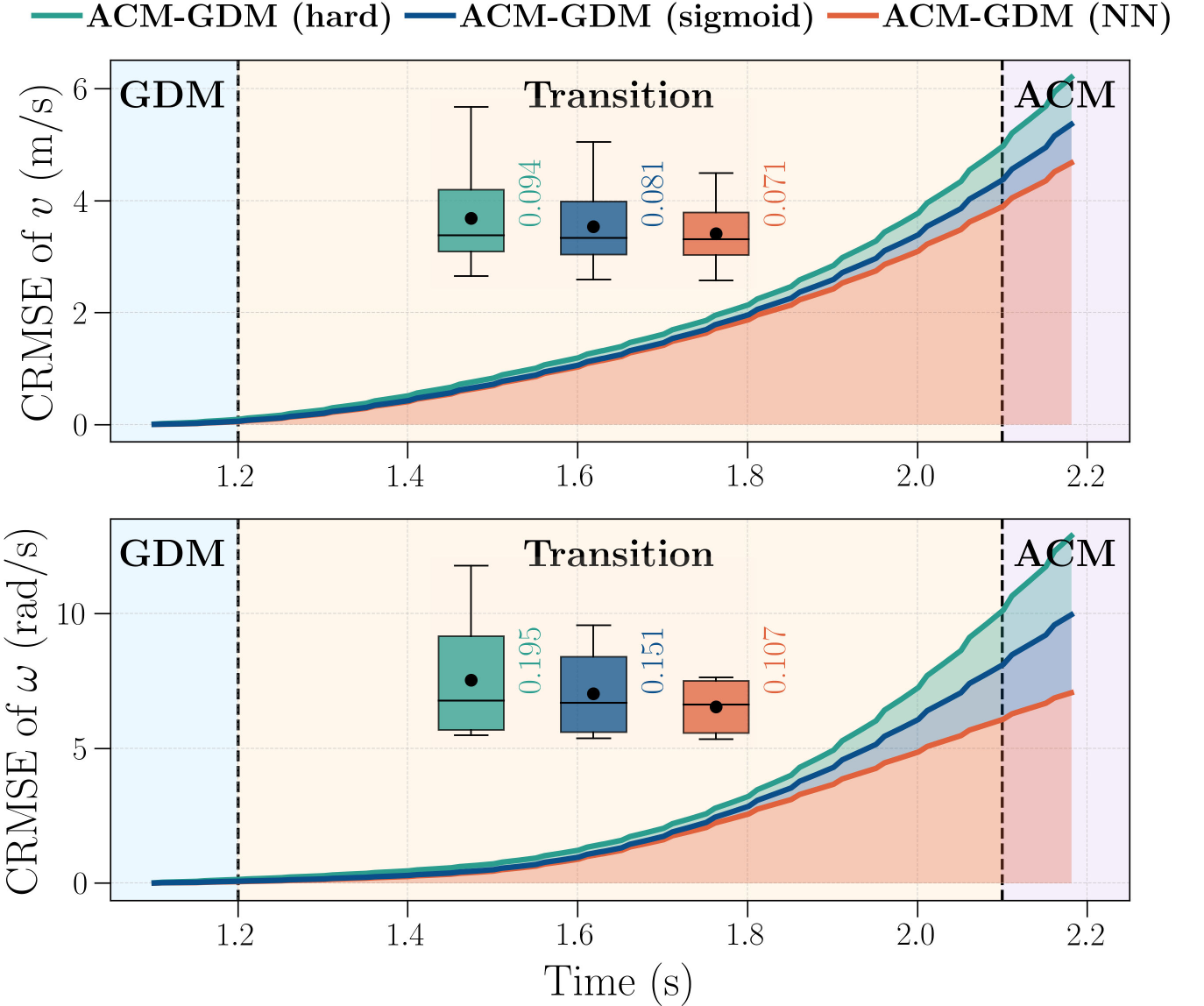}}
\caption{Cumulative RMSE comparison of translational and angular velocity predictions during a transition maneuver with control input $\left( F_l,F_r,\Delta\bar{r}_x \right)  = (7.57\,\text{gf},6.05\,\text{gf},4\,\text{cm})$, illustrating flight evolution from the GDM region to the ACM region. Results are presented for the ACM–GDM (hard), ACM–GDM (sigmoid) and ACM–GDM (NN) models, with inset boxplots summarizing RMSE distributions.}
\label{fig-CRMSE-mid}
\end{figure}

To illustrate, consider the example input $\left( F_l,F_r,\Delta\bar{r}_x \right) = (7.57\,\text{gf},6.05\,\text{gf},4\,\text{cm})$. Under this configuration, the blimp accelerates from a low-$V$, high-$\alpha$ GDM state into a high-$V$, low-$\alpha$ ACM state. The recorded sequence shows that the blimp remains in the GDM region until 1.2\,s, undergoes the transition from 1.2\,s to 2.1\,s, and enters the ACM region thereafter. Figure \ref{fig-CRMSE-mid} presents cumulative prediction errors for linear and angular velocities during this maneuver. We observe that ACM–GDM (hard) exhibits the largest cumulative error, ACM–GDM (sigmoid) shows intermediate error, and ACM–GDM (NN) achieves the smallest error. Quantitatively, for linear velocity, the average RMSE of ACM–GDM (NN) is reduced by 24.5\% compared to ACM–GDM (hard), and by 12.3\% compared to ACM–GDM (sigmoid). For angular velocity, the reductions are 45.1\% and 29.1\%, respectively. These results demonstrate that the learned, physically regularized transition produces more accurate and robust predictions during regime changes than either hard switching or a fixed sigmoid blend.

\subsection{Discussion}
Overall, the proposed ACM–GDM (NN) model effectively combines the complementary strengths of the two submodels across the entire flight envelope. Experimental results show that a single ACM fails to capture low-$V$, high-$\alpha$ behavior, while a single GDM performs poorly in high-$V$, low-$\alpha$ conditions. The hybrid formulation restores appropriate physics in both regimes by activating ACM where wing-generated lift–drag coupling is significant and GDM where bluff-body, viscous-dominated effects prevail. Quantitatively, as presented in Table~\ref{tab-RMSE}, ACM–GDM (NN) reduces the total average prediction loss on the full test set by 44.5\% relative to ACM-only and by 64.3\% relative to GDM-only. The learned transition further improves performance. ACM–GDM (NN) lowers total loss by 11.3\% versus the hard-switch hybrid and by 3.5\% versus the fixed-sigmoid hybrid, demonstrating that a data-driven, physically regularized blending mechanism yields more accurate predictions in the transition band without compromising consistency.

In short, the ACM–GDM (NN) model delivers an interpretable, physically grounded, and empirically validated aerodynamic model for winged blimps, providing accurate predictions across a wide range of speeds and angles of attack while maintaining smooth and physically plausible transitions between regimes.

\section{Conclusion}\label{section-Conclusion}
This paper presented an ACM–GDM hybrid model for the winged blimp, which unified a fixed-wing-style aerodynamic coupling model (ACM) and a blimp-style generalized drag model (GDM) via a learned, physically regularized neural-network transition. The hybrid model effectively captured distinct flight regimes defined by angle of attack and speed, with ACM dominating at higher $V$ and smaller $\alpha$, and GDM governing at lower $V$ and larger $\alpha$. The neural-network mixer provided a smooth and continuous interpolation across the transition band. Leveraging a three-phase identification pipeline and an extensive experimental campaign, the proposed hybrid model demonstrated accurate, interpretable predictions across the flight envelope and consistently outperformed single-model baselines as well as hard-switch or fixed-sigmoid blending strategies.

For future work, we plan to integrate the ACM–GDM hybrid model into closed-loop flight control, enabling stable and efficient operation of the winged blimp across a wider range of unstructured and dynamic flight conditions.

\bibliographystyle{IEEEtran}
\bibliography{IEEEabrv,paper}

\begin{thebibliography}{10}
\providecommand{\url}[1]{#1}
\csname url@samestyle\endcsname
\providecommand{\newblock}{\relax}
\providecommand{\bibinfo}[2]{#2}
\providecommand{\BIBentrySTDinterwordspacing}{\spaceskip=0pt\relax}
\providecommand{\BIBentryALTinterwordstretchfactor}{4}
\providecommand{\BIBentryALTinterwordspacing}{\spaceskip=\fontdimen2\font plus
\BIBentryALTinterwordstretchfactor\fontdimen3\font minus \fontdimen4\font\relax}
\providecommand{\BIBforeignlanguage}[2]{{%
\expandafter\ifx\csname l@#1\endcsname\relax
\typeout{** WARNING: IEEEtran.bst: No hyphenation pattern has been}%
\typeout{** loaded for the language `#1'. Using the pattern for}%
\typeout{** the default language instead.}%
\else
\language=\csname l@#1\endcsname
\fi
#2}}
\providecommand{\BIBdecl}{\relax}
\BIBdecl

\bibitem{Tmech-2021-swing-reducing}
Q.~Tao, J.~Wang, Z.~Xu, T.~X. Lin, Y.~Yuan, and F.~Zhang, ``Swing-reducing flight control system for an underactuated indoor miniature autonomous blimp,'' \emph{IEEE/ASME Transactions on Mechatronics}, vol.~26, no.~4, pp. 1895--1904, 2021.

\bibitem{Tmech-2021-scloud}
S.~H. Song, G.~Y. Yeon, H.~W. Shon, and H.~R. Choi, ``Design and control of soft unmanned aerial vehicle “s-cloud”,'' \emph{IEEE/ASME Transactions on Mechatronics}, vol.~26, no.~1, pp. 267--275, 2021.

\bibitem{TRO-RGBlimp-Q}
H.~Cheng and F.~Zhang, ``Rgblimp-q: Robotic gliding blimp with moving mass control based on a bird-inspired continuum arm,'' \emph{IEEE Transactions on Robotics}, vol.~41, pp. 5097--5116, 2025.

\bibitem{RAL-2018-sense}
V.~Mai, M.~Kamel, M.~Krebs, A.~Schaffner, D.~Meier, L.~Paull, and R.~Siegwart, ``Local positioning system using uwb range measurements for an unmanned blimp,'' \emph{IEEE Robotics and Automation Letters}, vol.~3, no.~4, pp. 2971--2978, 2018.

\bibitem{RAL-env-sense-2024}
S.~Sharma, M.~Verhoeff, F.~Joosen, R.~Venkatesha~Prasad, and S.~Hamaza, ``A morphing quadrotor-blimp with balloon failure resilience for mobile ecological sensing,'' \emph{IEEE Robotics and Automation Letters}, vol.~9, no.~7, pp. 6408--6415, 2024.

\bibitem{IROS-2017-interact}
N.~Yao, E.~Anaya, Q.~Tao, S.~Cho, H.~Zheng, and F.~Zhang, ``Monocular vision-based human following on miniature robotic blimp,'' in \emph{2017 IEEE International Conference on Robotics and Automation (ICRA)}, 2017, pp. 3244--3249.

\bibitem{ICRA-2019-model}
Y.~Wang, G.~Zheng, D.~Efimov, and W.~Perruquetti, ``Disturbance compensation based control for an indoor blimp robot,'' in \emph{2019 International Conference on Robotics and Automation (ICRA)}, 2019, pp. 2040--2046.

\bibitem{Blimp-review-2024}
\BIBentryALTinterwordspacing
S.~S. Bhat, S.~G. Anavatti, M.~Garratt, and S.~Ravi, ``Review of autonomous outdoor blimps and their applications,'' \emph{Drone Systems and Applications}, vol.~12, pp. 1--21, 2024. [Online]. Available: \url{https://doi.org/10.1139/dsa-2023-0052}
\BIBentrySTDinterwordspacing

\bibitem{RGBlimp-RAL}
H.~Cheng, Z.~Sha, Y.~Zhu, and F.~Zhang, ``Rgblimp: Robotic gliding blimp - design, modeling, development, and aerodynamics analysis,'' \emph{IEEE Robotics and Automation Letters}, vol.~8, no.~11, pp. 7273--7280, 2023.

\bibitem{Fixed_wing_fig_AIAA}
B.~M. Simmons, J.~L. Gresham, and C.~A. Woolsey, ``Flight-test system identification techniques and applications for small, low-cost, fixed-wing aircraft,'' \emph{Journal of Aircraft}, vol.~60, no.~5, pp. 1503--1521, 2023.

\bibitem{2013-ACM-book}
W.~Durham, \emph{Aircraft flight dynamics and control}.\hskip 1em plus 0.5em minus 0.4em\relax John Wiley \& Sons, 2013.

\bibitem{fixed-wing-TIM-2021}
Y.~Song, Z.~Liu, J.~Han, W.~Zhao, and N.~Li, ``Coordinated turn of fixed-wing aircraft under wind interface on integral backstepping,'' \emph{IEEE Transactions on Instrumentation and Measurement}, vol.~70, pp. 1--18, 2021.

\bibitem{fluid-2023-Re}
\BIBentryALTinterwordspacing
A.~Saez, M.~Manzo, and M.~Ciarcià, ``Numerical analysis of cambered plate configurations under low reynolds numbers and at a low-density condition,'' \emph{Fluids}, vol.~8, no.~7, 2023. [Online]. Available: \url{https://www.mdpi.com/2311-5521/8/7/194}
\BIBentrySTDinterwordspacing

\bibitem{Re-2023-aerospace}
\BIBentryALTinterwordspacing
C.~Robb and R.~Paul, ``Aerodynamic modeling of a flying wing featuring ludwig prandtl’s bell spanload,'' \emph{Aerospace}, vol.~10, no.~7, 2023. [Online]. Available: \url{https://www.mdpi.com/2226-4310/10/7/613}
\BIBentrySTDinterwordspacing

\bibitem{NSTSMC-2023-model-likeAUV}
F.~Mazzei, L.~Teofili, F.~Curti, and C.~Gargiulo, ``Mission analysis, dynamics and robust control of an indoor blimp in a cern detector magnetic environment,'' \emph{Frontiers in Robotics and AI}, vol.~10, 10 2023.

\bibitem{ICARM-2024-blimp-drag-control}
X.~Wang, J.~Zhang, Q.~Lu, H.~Li, S.~Hu, and A.~Song, ``Modeling and control for close flight formation of blimp robots,'' in \emph{2024 International Conference on Advanced Robotics and Mechatronics (ICARM)}, 2024, pp. 327--332.

\bibitem{fossen}
T.~I. Fossen, \emph{Nonlinear modelling and control of underwater vehicles}.\hskip 1em plus 0.5em minus 0.4em\relax Universitetet i Trondheim (Norway), 1991.

\bibitem{ACC-model-likeAUV}
J.~Wang and F.~Zhang, ``Achieving and maintaining inverted pose for miniature autonomous blimps,'' in \emph{2024 American Control Conference (ACC)}, 2024, pp. 338--343.

\bibitem{CCC-2023-model-likeAUV}
J.~Dong and Y.~Fang, ``Dynamics modeling and controller design for a novel indoor blimp,'' in \emph{2023 42nd Chinese Control Conference (CCC)}, 2023, pp. 298--303.

\bibitem{SBlimp-IROS-2023-control}
J.~Xu, D.~S. D'Antonio, D.~J. Ammirato, and D.~Saldaña, ``Sblimp: Design, model, and translational motion control for a swing-blimp,'' in \emph{2023 IEEE/RSJ International Conference on Intelligent Robots and Systems (IROS)}, 2023, pp. 6977--6982.

\bibitem{ICSL-2024-blimp-model-control}
M.~Kasmalkar, L.~Baird, and S.~Coogan, ``Feedback linearization of an underactuated miniature blimp with zero dynamics mitigation using high order control barrier functions,'' \emph{IEEE Control Systems Letters}, vol.~8, pp. 2589--2594, 2024.

\bibitem{stall-2015}
\BIBentryALTinterwordspacing
T.~T. Bui, ``\BIBforeignlanguage{en}{Analysis of stall aerodynamics of a swept wing with laminar-flow glove},'' \emph{\BIBforeignlanguage{en}{Journal of Aircraft}}, vol.~52, no.~3, p. 867–871, May 2015. [Online]. Available: \url{http://dx.doi.org/10.2514/1.c032883}
\BIBentrySTDinterwordspacing

\bibitem{high-alpha-2021}
\BIBentryALTinterwordspacing
M.~Manolesos and G.~Papadakis, ``\BIBforeignlanguage{en}{Investigation of the three-dimensional flow past a flatback wind turbine airfoil at high angles of attack},'' \emph{\BIBforeignlanguage{en}{Physics of Fluids}}, vol.~33, no.~8, Aug 2021. [Online]. Available: \url{http://dx.doi.org/10.1063/5.0055822}
\BIBentrySTDinterwordspacing

\bibitem{bluff-body-2022}
\BIBentryALTinterwordspacing
Z.~Zhang, Z.~Wang, and I.~Gursul, ``\BIBforeignlanguage{en}{Aerodynamics of a wing in turbulent bluff body wakes},'' \emph{\BIBforeignlanguage{en}{Journal of Fluid Mechanics}}, vol. 937, Mar 2022. [Online]. Available: \url{http://dx.doi.org/10.1017/jfm.2022.132}
\BIBentrySTDinterwordspacing

\bibitem{Neural-ODE}
R.~T. Chen, Y.~Rubanova, J.~Bettencourt, and D.~K. Duvenaud, ``Neural ordinary differential equations,'' \emph{Advances in neural information processing systems}, vol.~31, 2018.

\bibitem{ABNODE}
Y.~Zhu, H.~Cheng, and F.~Zhang, ``Data-driven dynamics modeling of miniature robotic blimps using neural odes with parameter auto-tuning,'' \emph{IEEE Robotics and Automation Letters}, vol.~9, no.~12, pp. 10\,986--10\,993, 2024.

\bibitem{2022-Re}
\BIBentryALTinterwordspacing
E.~Balla and J.~Vad, ``\BIBforeignlanguage{en}{Models for estimation of lift and drag coefficients for low-reynolds-number cambered plates},'' \emph{\BIBforeignlanguage{en}{AIAA Journal}}, vol.~60, no.~12, p. 6620–6632, Dec 2022. [Online]. Available: \url{http://dx.doi.org/10.2514/1.J061579}
\BIBentrySTDinterwordspacing

\bibitem{aerospace-2024-Re}
\BIBentryALTinterwordspacing
A.~Arshad and V.~Kovaļčuks, ``Computational investigations for the feasibility of passive flow control devices for enhanced aerodynamics of small-scale uavs,'' \emph{Aerospace}, vol.~11, no.~6, 2024. [Online]. Available: \url{https://www.mdpi.com/2226-4310/11/6/473}
\BIBentrySTDinterwordspacing

\bibitem{TRO-2024-UKF}
A.~Saviolo, J.~Frey, A.~Rathod, M.~Diehl, and G.~Loianno, ``Active learning of discrete-time dynamics for uncertainty-aware model predictive control,'' \emph{IEEE Transactions on Robotics}, vol.~40, pp. 1273--1291, 2024.

\bibitem{KNODE-MPC-2022}
K.~Y. Chee, T.~Z. Jiahao, and M.~A. Hsieh, ``Knode-mpc: A knowledge-based data-driven predictive control framework for aerial robots,'' \emph{IEEE Robotics and Automation Letters}, vol.~7, no.~2, pp. 2819--2826, 2022.

\bibitem{stall-2019}
\BIBentryALTinterwordspacing
D.~Liu and T.~Nishino, ``\BIBforeignlanguage{en}{Unsteady rans simulations of strong and weak 3d stall cells on a 2d pitching aerofoil},'' \emph{\BIBforeignlanguage{en}{Fluids}}, vol.~4, no.~1, p.~40, Mar 2019. [Online]. Available: \url{http://dx.doi.org/10.3390/fluids4010040}
\BIBentrySTDinterwordspacing

\bibitem{fluids-2023-critical-Re}
\BIBentryALTinterwordspacing
A.~Utkina, A.~Kozelkov, R.~Zhuchkov, and D.~Strelets, ``Numerical study of the influence of the critical reynolds number on the aerodynamic characteristics of the wing airfoil,'' \emph{Fluids}, vol.~8, no.~10, 2023. [Online]. Available: \url{https://www.mdpi.com/2311-5521/8/10/276}
\BIBentrySTDinterwordspacing

\end{thebibliography}

\raggedbottom 

\begin{IEEEbiography}[{\includegraphics[width=1in,height=1.25in,clip,keepaspectratio]{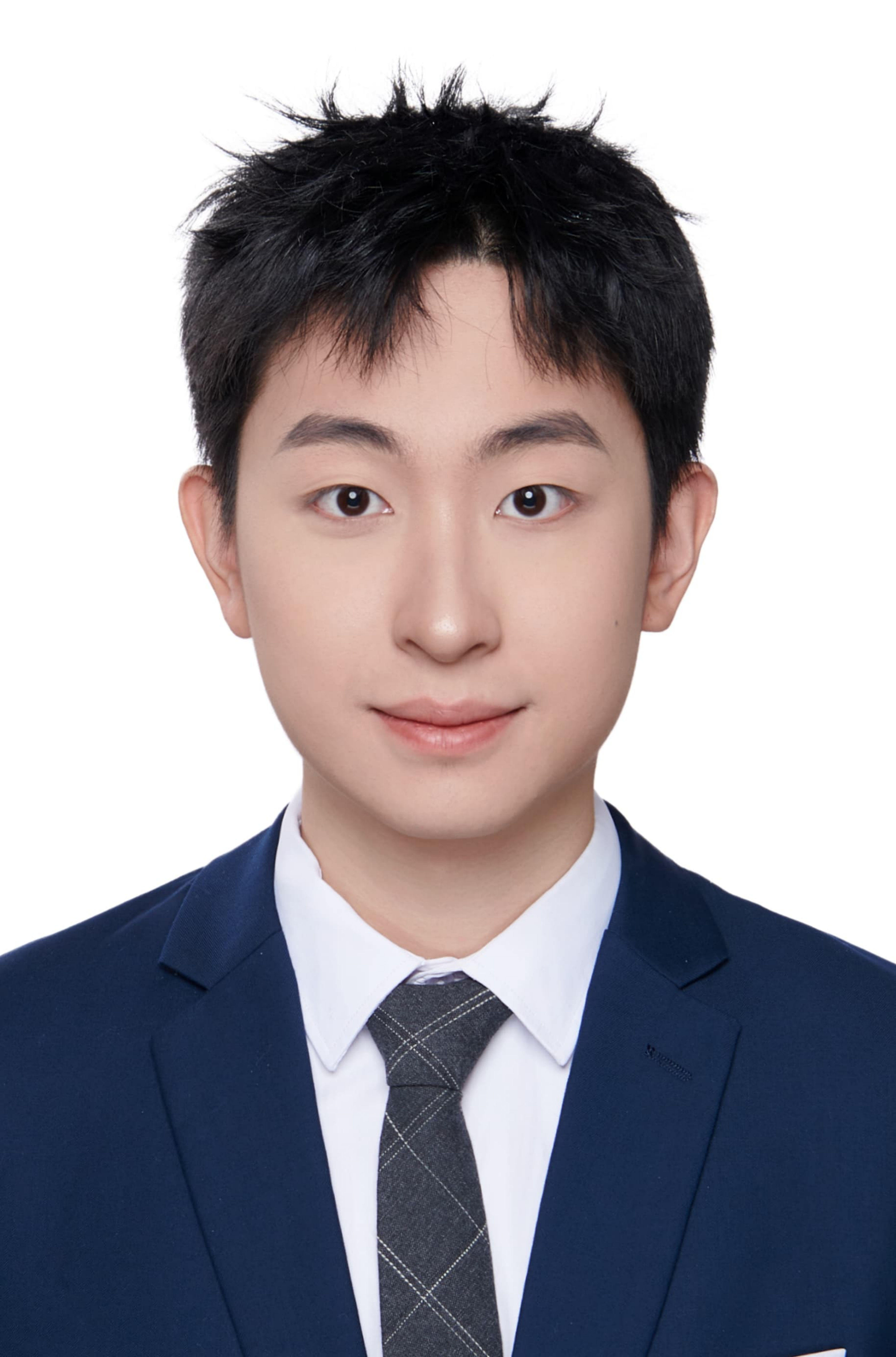}}]{Xiaorui Wang} (Student Member, IEEE) 
received the bachelor’s degree in robotics engineering in 2025 from Peking University, Beijing, China, where he is currently working toward the Ph.D. degree in general mechanics and foundation of mechanics with the School of Advanced Manufacturing and Robotics, Peking University, Beijing, China. 

His research interests include aerial vehicles, dynamical modeling, and learning-based control.
\end{IEEEbiography}

\begin{IEEEbiography}[{\includegraphics[width=1in,height=1.25in,clip,keepaspectratio]{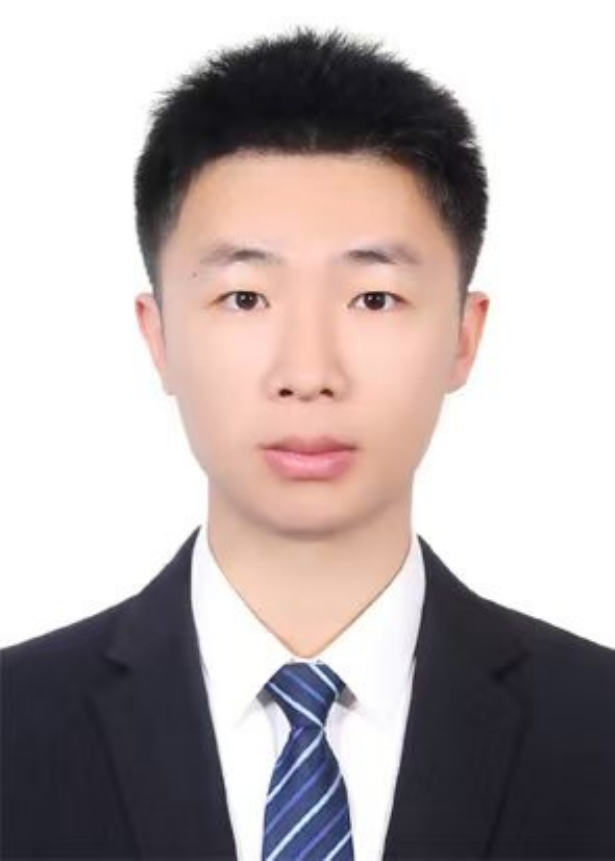}}]{Hongwu Wang} 
received the bachelor’s degree in Robotics Engineering from Harbin Institute of Technology, Harbin, China, in 2024. He is currently working toward the M.S. degree in mechanical engineering with the School of Advanced Manufacturing and Robotics, Peking University, Beijing, China.  

His research interests include aerial vehicles, mechatronics
systems, reinforcement learning and robotic control.
\end{IEEEbiography}

\begin{IEEEbiography}[{\includegraphics[width=1in,height=1.25in,clip,keepaspectratio]{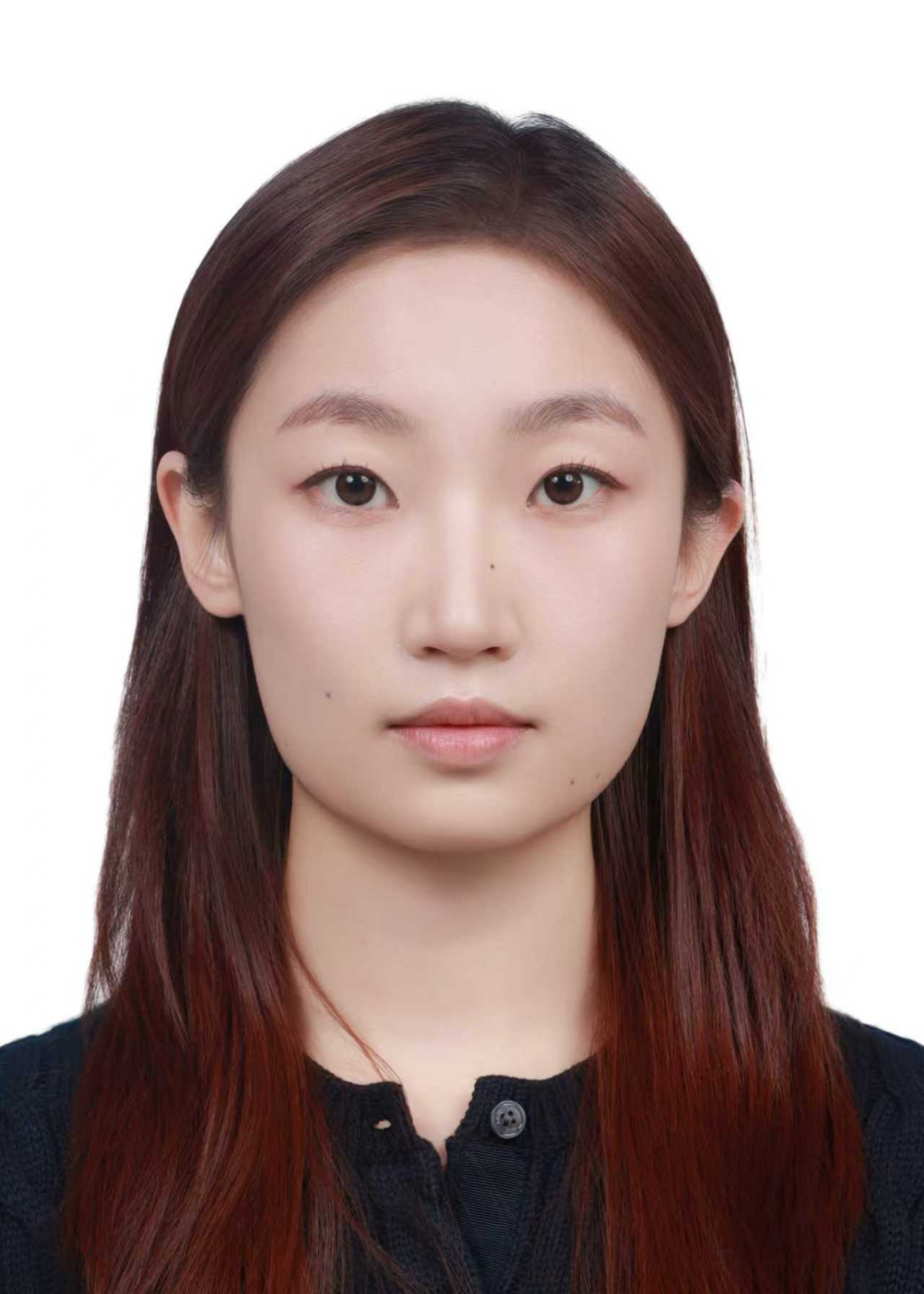}}]{Yue Fan} 
received the B.S. degree in Software Engineering from China University of Geosciences (Beijing), Beijing, China, in 2023. She is currently working toward the M.S. degree in mechanical engineering with the School of Advanced Manufacturing and Robotics, Peking University, Beijing, China. 

Her research interests include aerial vehicles, deep learning, and robot learning.
\end{IEEEbiography}

\begin{IEEEbiography}[{\includegraphics[width=1in,height=1.25in,clip,keepaspectratio]{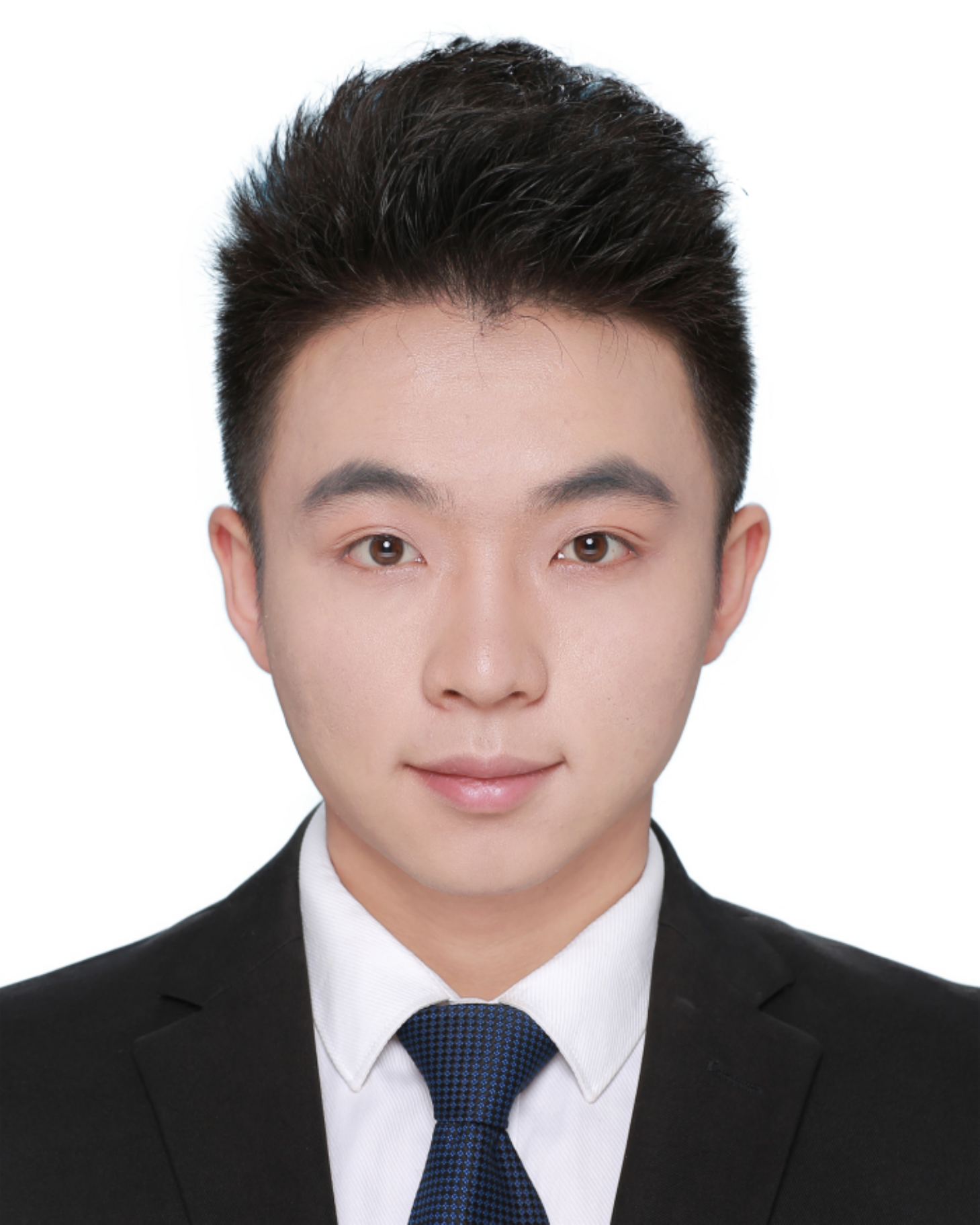}}]{Hao Cheng} (Student Member, IEEE) 
received the Bachelor's degree in New Energy Science and Engineering (Wind Power) from North China Electric Power University, China, in 2017, and the Master's degree in Control Engineering from Tsinghua University, China, in 2021. 
He is currently working toward the Ph.D. degree in general mechanics and foundation of mechanics with the School of Advanced Manufacturing and Robotics, Peking University, Beijing, China. 

His research interests include \mbox{lighter-than-air} aerial vehicles, bio-inspired robotics, and continuum robots. 
\end{IEEEbiography}

\begin{IEEEbiography}[{\includegraphics[width=1in,height=1.25in,clip,keepaspectratio]{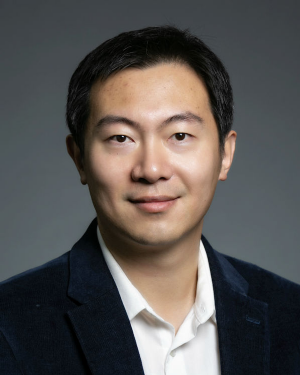}}]{Feitian Zhang} (Member, IEEE) 
received the bachelor’s and master’s degrees in automatic control from the Harbin Institute of Technology, Harbin, China, in 2007 and 2009, respectively, and the Ph.D. degree in electrical and computer engineering from Michigan State University, East Lansing, MI, USA, in 2014.

He was a Postdoctoral Research Associate with the Department of Aerospace Engineering and Institute for Systems Research, University of Maryland, College Park, MD, USA, from 2014 to 2016, and an Assistant Professor of Electrical and Computer Engineering with George Mason University, Fairfax, VA, USA, from 2016 to 2021. He is currently an Associate Professor of Robotics Engineering with Peking University, Beijing, China. His research interests include mechatronics systems, robotics and controls, aerial vehicles, and underwater vehicles. 
\end{IEEEbiography}

\end{document}